\documentclass{article}

\usepackage{arxiv}

\usepackage{amsmath}
\usepackage{bm}
\usepackage{amsbsy}
\usepackage{amsfonts}
\usepackage{setspace}
\usepackage{times}
\usepackage{lineno}
\usepackage{import}

\usepackage{multirow,bigdelim}

\usepackage{graphicx}
\usepackage{xcolor}
\usepackage{caption}
\usepackage{subcaption}
\usepackage{url}
\usepackage{booktabs}

\usepackage{textcomp}
\usepackage{gensymb}

\usepackage{siunitx}

\usepackage[colorlinks=false, 
            citebordercolor=blue, 
            linkbordercolor=red,  
            urlbordercolor=green, 
            breaklinks=true,      
            hidelinks]{hyperref}

\newcommand{\eg}{\textit{e.g.},~}
\newcommand{\ie}{\textit{i.e.},~}
\newcommand{\Eqn}[1]{Eqn.~\ref{#1}}

\newcommand{\Fig}[1]{Fig.~\ref{#1}}
\newcommand{\Figu}[1]{Figure~\ref{#1}}
\newcommand{\Figs}[2]{Figs.~\ref{#1} and \ref{#2}}
\newcommand{\Figz}[2]{Fig.~\ref{#1}--\ref{#2}}

\newcommand{\Tab}[1]{Table~\ref{#1}}
\newcommand{\Sec}[1]{Section~\ref{#1}}

\newcommand{\uu}{{\mathbf u }}

\newcommand{\sm}{{\mathbf s }}

\newcommand{\thetaa}{\pmb{\theta}}
 
\newcommand{\chii}{\pmb{\chi}}

\title{Using Deep Operators to Create Spatio-temporal Surrogates for Dynamical Systems under Uncertainty}

\author{ 
{Jichuan Tang}
  \\
	Department of Civil and Environmental\\  Engineering and Earth Sciences\\
	University of Notre Dame\\
	Notre Dame, IN 46556 \\
	\texttt{jtang4@nd.edu} \\
 \And 
 {Patrick T. Brewick} \\
	Department of Civil and Environmental\\  Engineering and Earth Sciences\\
	University of Notre Dame\\
	Notre Dame, IN 46556 \\
	\texttt{pbrewick@nd.edu} \\ 
	\And
 {Ryan G. McClarren} \\
	Department of Aerospace and Mechanical Engineering\\
	University of Notre Dame\\
	Notre Dame, IN 46556 \\
	\And
 {Christopher Sweet} \\
	Center for Research Computing\\
	University of Notre Dame\\
	Notre Dame, IN 46556 \\
}

\renewcommand{\shorttitle}{Creating Spatio-Temporal DeepONets for Dynamical Systems}

\begin{document}

\maketitle

\begin{abstract}
Spatio-temporal data, which consists of responses or measurements gathered at different times and positions, is ubiquitous across diverse applications of civil infrastructure.  While SciML methods have made significant progress in tackling the issue of response prediction for individual time histories, creating a full spatial-temporal surrogate remains a challenge.  This study proposes a novel variant of deep operator networks (DeepONets), namely the full-field Extended DeepONet (FExD), to serve as a spatial-temporal surrogate that provides multi-output response predictions for dynamical systems. The proposed FExD surrogate model effectively learns the full solution operator across multiple degrees of freedom by enhancing the expressiveness of the branch network and expanding the predictive capabilities of the trunk network. The proposed FExD surrogate is deployed to simultaneously capture the dynamics at several sensing locations along a testbed model of a cable-stayed bridge subjected to stochastic ground motions. The ensuing response predictions from the FExD are comprehensively compared against both a vanilla DeepONet and a modified spatio-temporal Extended DeepONet.  The results demonstrate the proposed FExD can achieve both superior accuracy and computational efficiency, representing a significant advancement in operator learning for structural dynamics applications.
\end{abstract}


% \linenumbers
\maketitle

\section{Introduction} \label{sec1}

Modeling complex dynamical systems remains a challenging task for several reasons, ranging from the presence of spatial-temporal relationships to the potential high-dimensionality of the model and/or its parameters. While advances in computational modeling, e.g., finite element (FE) models, have brought about dramatic increases in the accuracy and fidelity of full-field response predictions for dynamical systems, these gains are typically accompanied by high computational costs. Model-order reduction techniques can significantly decrease computational costs by creating low-dimensional surrogate models \cite{li2023improved,mallick2025aibased}. However, these models can incur high offline costs and may exhibit inadequate prediction accuracy. Another prevalent method is polynomial chaos \cite{harosandoval2012sensitivity}, which is employed for stochastic systems characterized by arbitrary random variables. Standard techniques such as Galerkin projection \cite{matthies2003nonlinear}, stochastic collocation \cite{hubler2020global}, and least squares approximation \cite{kaveh2024applications} are utilized to determine the polynomial chaos coefficients for surrogate models. However, when numerous realizations are required for objectives such as optimization or uncertainty quantification, the total computational costs can still become wholly impractical. The modeling challenge can become further intensified when attempting to address stochastic excitation sources, behavioral nonlinearities and noisy or incomplete measurements.

Fortunately, surrogate modeling methods that integrate emergent scientific machine learning (SciML) techniques are proving to effectively counter the computational burden associated with the complex physics-based models without having to greatly sacrifice accuracy \cite{samadian2025application,kontolati2024learning,ahmed2025physicsinformed,gao2025quantifying}. Utilizing SciML-based methods to predict spatial-temporal responses for dynamical systems has become an active area of research with several cutting-edge methods. Recent advances include latent-space neural operator learning approaches for facilitating real-time predictions for highly nonlinear and multiscale systems on high-dimensional domains\cite{kontolati2024learning}, long short-term memory networks (LSTMs) for sequential data \cite{dang2023vibrationbased}, graph neural networks (GNNs) for topology-aware predictions \cite{wen2023datadriven}, and physics-informed neural networks (PINNs) that encodes the prior physics knowledge in the network architecture \cite{rao2023encoding}.  In addition, a diffusion model for dynamics-informed diffusion processes has been proposed \cite{ruhlingcachay2023dyffusion}, as have generative models to reconstruct and predict full-field spatio-temporal dynamics on the basis of sparse measurements \cite{li2024learning} and a Bayesian neural field \cite{saad2024scalable} as a domain-general statistical model that infers rich spatio-temporal probability distributions for data analysis tasks. Among this diversity of approaches, operator-based learning has also emerged as a popular surrogate for a variety of systems given its basis in the universal approximation theorem \cite{tianpingchen1995universal}. This makes deep operator networks (DeepONets) especially well-suited for capturing dynamical systems, as they not only learn the mapping relationship between functions but also try to learn the underlying dynamical systems and solve the associated partial differential equations (PDEs) \cite{lu2021learning}. Another notable advantage of the DeepONet is its flexible architecture, which has inspired several variants of neural operators, namely the Fourier neural operator (FNO)\cite{kaewnuratchadasorn2024neural}, Laplace neural operator (LNO)\cite{cao2024laplace}, and causality-DeepONet \cite{liu2022causalitydeeponet}, among others. 

Building upon these advantages, the standard DeepONet (called the ``vanilla'' DeepONet in this paper) and its variants have demonstrated great potential for applications in structural engineering, such as structural systems exposed to natural hazards \cite{liu2022causalitydeeponet, goswami2025neural}, bi-fidelity modeling for hysteretic systems under uncertainty \cite{de2024bifidelity}, floating offshore structures under irregular waves \cite{cao2024deep}, and ground settlement in real-time during mechanized tunnel excavation \cite{xu2024multifidelity}, among other examples \cite{pickering2022discovering,garg2022assessment}. While the results from these studies are promising, the applications have so far been limited to learning responses from individual degrees of freedom. % because DeepONets face challenges in generalizing spatial information and extrapolating beyond training regimes. 
To this end, Lu et al. have proposed several approaches for multiple output functions \cite{lu2022comprehensive}. For instance, one suggested method is to use multiple independent DeepONets with each DeepONet providing only one function \cite{garg2022assessment}; however, this approach is not always highly efficient as the number of outputs increases. Alternatively, a single DeepONet can be designed for multiple outputs by modifying the network architecture. Notably, this strategy enhances efficiency by leveraging shared network architectures, such as splitting and integrating multiple DeepONets into a unified framework to predict multiple static structural responses \cite{ahmed2025physicsinformed}. In addition, He et al.~developed sequential DeepONets with LSTM and gated recurrent unit (GRU) networks for predicting full-field solutions under time-dependent loads \cite{he2024sequential} and introduced Geom-DeepONet to encode parameterized 3D geometries for predicting solutions across arbitrary node configurations \cite{he2024geomdeeponet}.

However, investigation and comparison between these multi-output approaches, both theoretically and computationally, remain limited on how to effectively capture the spatial correlations inherent in structural dynamics problems, where strong interdependencies exist among different degrees of freedom. This gap is particularly evident when considering applications involving complex spatio-temporal dynamics, where traditional DeepONet architectures may struggle to efficiently model the full solution field. As pointed out by Li et al.~\cite{li2025architectural}, the capability of the DeepONet for approximate solutions to nonlinear PDEs might be constrained by its architecture, which relies primarily on basis expansion through the dot product of sub-network outputs. To address this limitation, Li et al.~proposed an extended DeepONet (Ex-DeepONet) that allows for nonlinear interaction between the branch and trunk networks, moving beyond the linear combinations of conventional DeepONet \cite{li2025architectural}. Similarly, Haghighat et al.~\cite{haghighat2024endeeponet} introduced the Enriched-DeepONet (En-DeepONet) to address the limitations of current operator learning models in dealing with moving-solution operators.

This study introduces the full-field extended DeepONet to learn the full spatial-temporal solution mesh for a given dynamical system through novel modifications to the underlying operator network architecture. The full-field Ex-DeepONet architecture improves computational accuracy and efficiency and advances operator learning-based surrogate modeling for spatial-temporal prediction.  The full-field Ex-DeepONet extends the enrichment approach to enhance the expressivity of neural operators by hierarchically combining information from branch networks to yield simultaneous predictions of dynamical responses at multiple degrees-of-freedom (DoFs).  Such an approach is beneficial for capturing spatial correlations inherent in dynamical systems, improving network accuracy for systems with strong inter-DoF dependencies. The effectiveness of the proposed full-field Ex-DeepONet is demonstrated through a detailed structural dynamics application. By addressing the limitations of existing multi-output strategies, the proposed approach achieves superior performance in capturing a range of dynamic responses from a complex structural system while also offering a computational efficiency advantage.
 
The remainder of this paper is organized as follows: Section 2 formulates the general problem of finding the structural operator for dynamical systems and outlines the preliminaries of surrogate modeling using the vanilla, Ex-DeepONet, and proposed full-field Ex-DeepONet for spatial-temporal predictions. Section 3 introduces the considered structural dynamics application of a bridge structure subjected to stochastic ground excitation. The data generation process, model architectures, and training hyper-parameters are also described. Section 4 evaluates and discusses the performance of the various considered models for dynamic response prediction. Finally, Section 5 summarizes the primary conclusions and contributions, along with presenting a future outlook. 

\section{Methodology}

This section first introduces operator networks and presents the vanilla DeepONet methodology for predicting time-series responses at single or multiple locations. We then introduce the Ex-DeepONet and full-field Ex-DeepONet, an extension that achieves multi-degree-of-freedom (MDOF) predictions by considering spatial-temporal coordinates and leveraging full-field vector-based methods.

\subsection{Operator for Dynamical System}

Consider a dynamical system characterized by a finite set of input-output pairs. Assume that the input functions belong to the Banach space $\mathcal{U} = \left\{ \Omega; u : \mathcal{X} \to \mathbb{R}^{d_u} \right\}$, where $\mathcal{X} \subset \mathbb{R}^{d_\chi}$, and let the corresponding output functions belong to $\mathcal{S} = \left\{ \Omega; s : \mathcal{Y} \to \mathbb{R}^{d_s} \right\}$, where $\mathcal{Y} \subset \mathbb{R}^{d_y}$, respectively.  The governing equation for a dynamical system can be expressed as the operator mapping between two infinite-dimensional spaces on a bounded open set \(\Omega \subset \mathbb{R}^D\), as presented in \Eqn{eq:gov} 
\begin{equation} \label{eq:gov}
    \mathcal{L}(s)(\chii) = u(\chii), \qquad \chii \in \Omega
\end{equation} 
where $\mathcal{L}$ denotes a differential operator; $\chii$ represents a location inside the domain $\Omega$; and $u$ is some external forcing function and $s$ is the response of the system. Assume the solution operator to \Eqn{eq:gov} is given by $\mathcal{G}(u)(\chii) = s(\chii)$. The goal of operator learning is to approximate the solution operator, \(\mathcal{G}\), with $\mathcal{G}_{\theta}$ as shown in \Eqn{eq:G_mapping}:
\begin{equation}
\mathcal{G} : \mathcal{U} \times \Theta \to \mathcal{S} \quad \text{or}  \quad \mathcal{G} \approx  \mathcal{G}_{\theta} : \mathcal{U} \to \mathcal{S}, \quad \theta \in \Theta, \label{eq:G_mapping}
\end{equation}
where \(\Theta\) is a finite-dimensional parameter space and $\mathcal{G}_{\theta}$ is an approximation of solution operator $\mathcal{G}$, namely a mapping from input functions space $\mathcal{U}$ to output functions space $\mathcal{S}$. In the standard setting, the optimal parameters \(\theta^*\) are learned by training the neural operator with a set of labeled observations \(\mathcal{D} = \left\{ (\uu^{(q)}, \sm^{(q)}) \right\}_{q=1}^{N_u}\), which contains \(N_u\) pairs of input and output functions $\uu^{(q)}$ and $\sm^{(q)}$, respectively. Notably, the physical system involves multiple functions, such as the solution, the forcing term, the initial condition, and the boundary conditions. We are typically interested in predicting one of these functions, which is the output of the solution operator (defined on the space \(\mathcal{S}\)), based on the varied forms of the other functions, namely, the input functions in the space \(\mathcal{U}\).

\subsection{Deep Operator Network (DeepONet)}\label{sec:deeponet}

\begin{figure}[tb]
	\centering
	\begin{subfigure}[b]{0.48\textwidth}
		\centering
        \includegraphics[width=\linewidth]{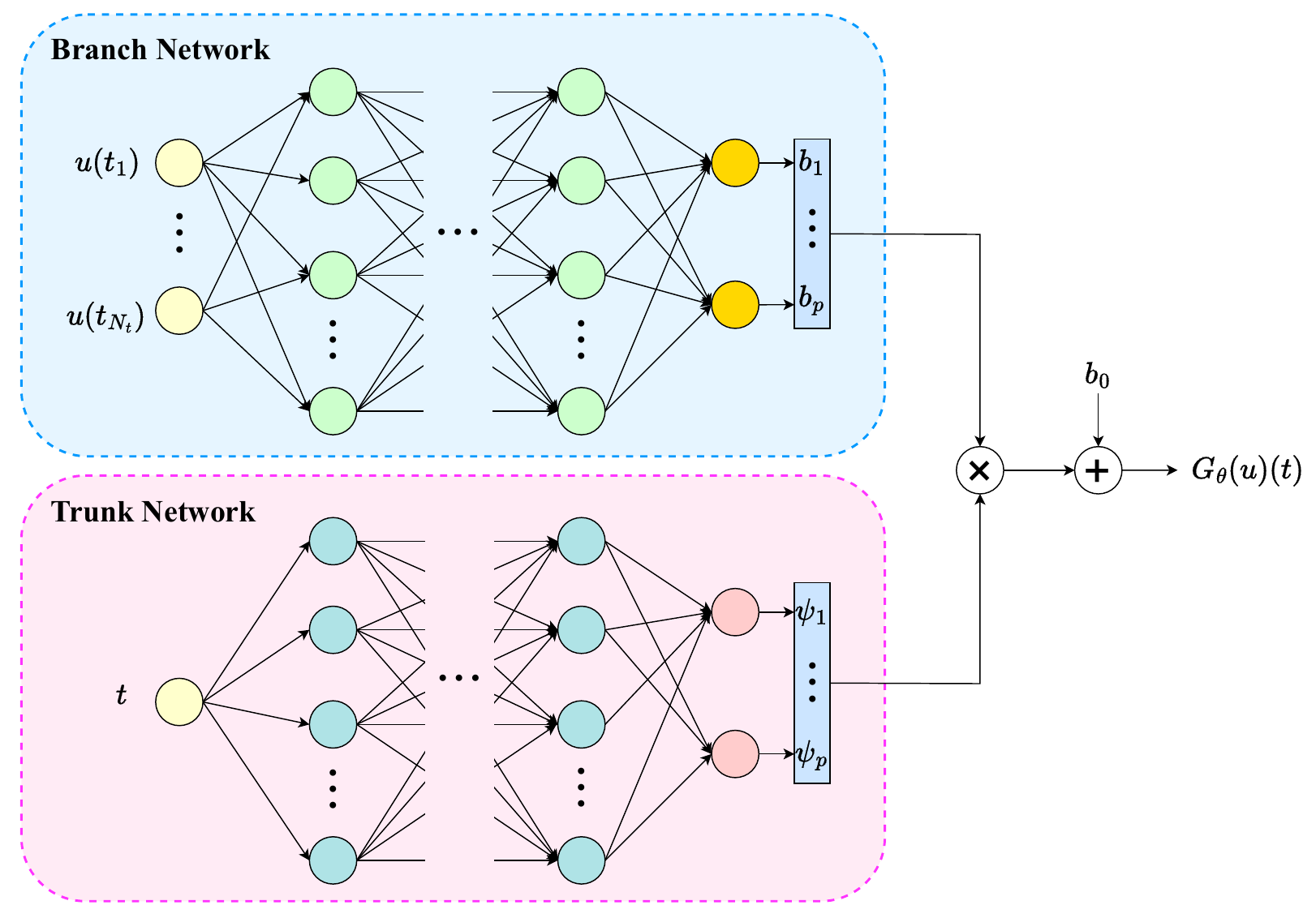}
        \caption{Vanilla DeepONet for SDOF response}\label{fig:single_output_vanilla_DON}
	\end{subfigure}
 \vspace{9pt}
     \begin{subfigure}[b]{0.48\textwidth}
		\centering
		\includegraphics[width=\linewidth]{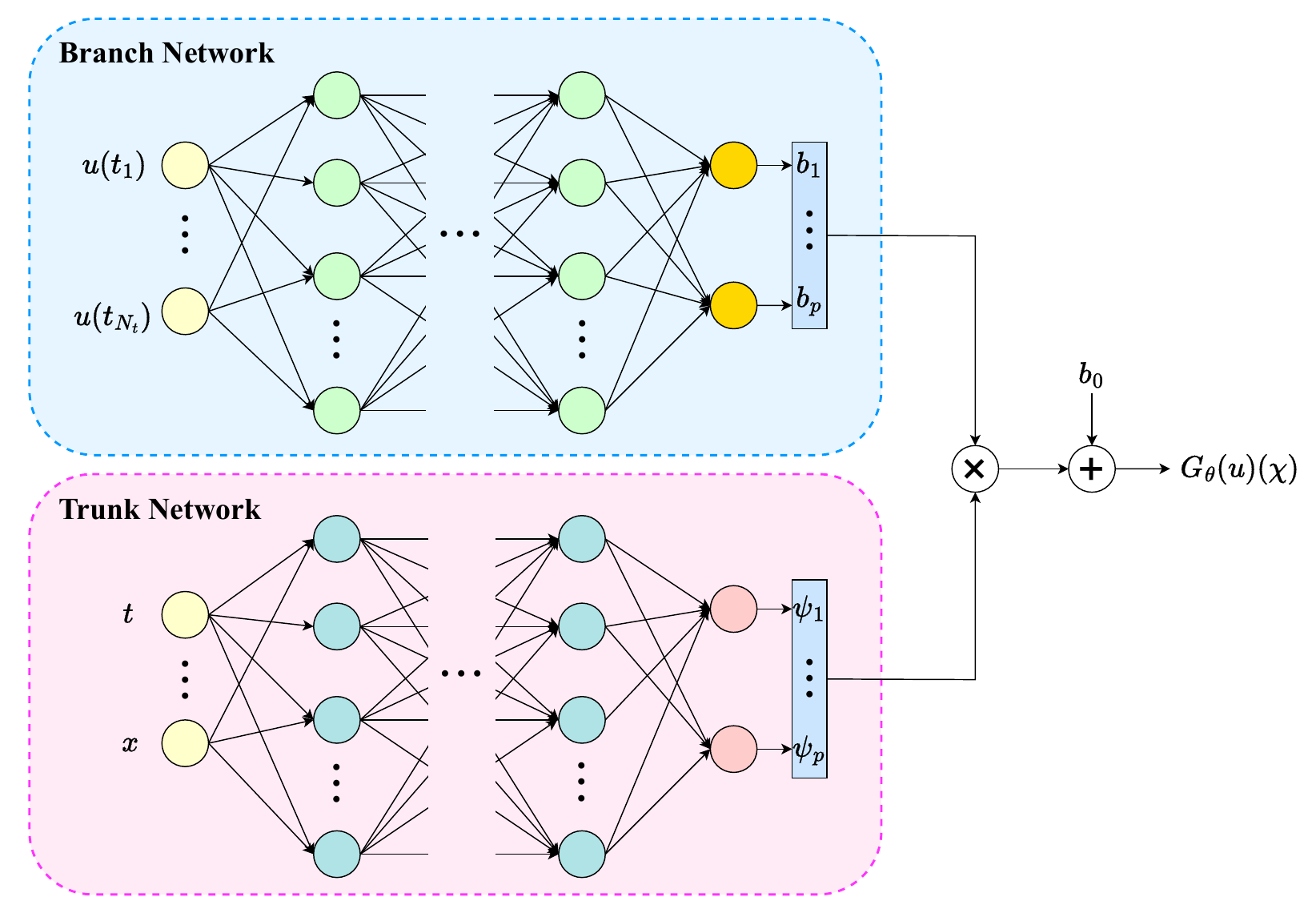}
        \caption{vanilla DeepONet for MDOF responses}\label{fig:multi_output_vanilla_DON}
	\end{subfigure}
	\caption{Schematic  illustration of the vanilla DeepONet model considering (a) individual DoF responses and (b) MDoF responses. 
    }\label{fig:vanilla_deeponet}  
\end{figure}

The DeepONet is based on the universal approximation theorem for operators \cite{chen1995universal}.  The architecture of the vanilla DeepONet is comprised of two deep neural networks: a \textit{branch network}, which encodes the input at specific sensing points, and the \textit{trunk network}, which encodes information about the coordinates at which the solution operator is evaluated \cite{lu2021learning}. For the case of time series problems within dynamical systems, the branch network can encode realizations of the input function discretized at fixed temporal coordinates as in $\{u(t_1),u(t_2),\ldots,u(t_{N_t})\}$ where $N_t$ represents the number of measurement or sensor locations. Note that $N_t$ becomes the number of time points in the case of a time series problem. Meanwhile, regarding the trunk network, previous studies using DeepONets for dynamical systems have only included the coordinate for time, namely $\chii = \{t_i | i = 1,\dots,N_t\}$ \cite{goswami2025neural,garg2022assessment}. With this construction, the solution operator is only designed to predict a single DoF response, which is not associated with different spatial information, as illustrated in \Fig{fig:single_output_vanilla_DON}. In such cases, multiple independent DeepONets are required to obtain a full spatial-temporal solution for an MDOF dynamical system. 

This study adjusts the trunk network to encode information related to the full spatial-temporal coordinates, as depicted in \Fig{fig:multi_output_vanilla_DON}, where the solution operator is evaluated at $\chi_{i,j} = (t_i, x_j)$ for $i=1,\dots,N_t$ and $j=1,\dots,N_x$ with $N_t$ and $N_x$ denoting the temporal and spatial discretization, respectively. For each fixed spatial location $x_j$, the set of all time instants $t_i$ forms the discrete spatio-temporal coordinates $\chii_j = \{\chi_{i,j}\mid i = 1, \ldots, N_t\}$ . The full spatio-temporal grid then comes from the union of all $\chii_j$, \ie $\chii = \bigcup_{j=1}^{N_x} \chii_j$. 

Both the branch and trunk networks are designed with the architecture of a multilayer perception (MLP), as schematically illustrated in \Fig{fig:vanilla_deeponet}. The branch network has $N_{l_b}$ hidden layers and a final output layer, \ie the $(N_{l_b} + 1)^{\text{th}}$ layer, that contains $p$ neurons. When an input function $\mathbf{u} = [(u(t_1), u(t_2), \ldots, u(t_{N_t})]^{\mathrm{T}}$ is given to the branch network, it produces an output feature $\mathbf{z}_b^l$ of each hidden layer $\ell = 1, 2, \ldots, N_{l_b}$. The eventual output from final layer ($N_{l_b} + 1$) is represented as $[b_1, b_2, \ldots, b_p]^{\mathrm{T}}$. The output features of each hidden layer $\ell = 1, 2, \ldots, N_{l_b}$ and final layer ($N_{l_b} + 1$) can be expressed as in \Eqn{eq:mlp_branch}
\begin{subequations}
\begin{align}
    \mathbf{z}_b^\ell &= \mathcal{R}_b^\ell (\mathbf{W}_b^\ell \mathbf{z}_b^{\ell-1} + \mathbf{B}_b^\ell)\\
    \mathbf{z}_b &= [b_1, b_2, \ldots, b_p]^{\mathrm{T}} = \mathcal{R}_b^{N_{l_b}+1} (\mathbf{W}_b^{N_{l_b}+1} \mathbf{z}_b^{N_{l_b}} + \mathbf{B}_b^{N_{l_b}+1})
\end{align}\label{eq:mlp_branch}
\end{subequations}
where $\mathbf{z}_b^0 = \uu^{(q)}$, denoting the initial input vector of branch network, which is the discrete measurements of the $q$th sample’s forcing function at temporal coordinates $\{t\}_{i=1}^{N_t}$. $\mathbf{W}_b^\ell$ and $\mathbf{B}_b^\ell$ are the weights and biases of the $\ell$-th layer of branch network, and $\mathcal{R}_b^\ell(\cdot)$ denotes the nonlinear activation function. 

Regarding the trunk network with $N_{l_t}$ hidden layers, it takes as input spatial-temporal coordinate $\chii_j$. The output features of each hidden layer $\ell = 1, 2, \ldots, N_{l_t}$ and final layer ($N_{l_t} + 1$) is shown in \Eqn{eq:mlp_trunk}.
\begin{subequations}
\begin{align}
    \mathbf{z}_t^\ell &= \mathcal{R}_t^\ell (\mathbf{W}_t^\ell \mathbf{z}_t^{\ell-1} + \mathbf{B}_t^\ell)\\
    \mathbf{\psi}_t &= [\psi_1, \psi_2, \ldots, \psi_p]^\mathrm{T} = \mathcal{R}_t^{N_{l_t}+1} (\mathbf{W}_t^{N_{l_t}+1} \mathbf{z}_t^{N_{l_t}} + \mathbf{B}_t^{N_{l_t}+1})
\end{align}\label{eq:mlp_trunk}
\end{subequations}
where the initial input for the trunk network denotes the set of spatio-temporal coordinates at a fixed spatial point $x_j$ across all temporal points $t_i$, \ie $\mathbf{z}_t^0 = \chii_j = \{(t_i, x_j)\mid i = 1, \ldots, N_t\}$, $\mathbf{W}_t^\ell$ and $\mathbf{B}_t^\ell$ are the weights and biases of the $\ell$-th layer of trunk network, respectively, and $\mathcal{R}_t^\ell(\cdot)$ is the nonlinear activation function.  Note that for traditional applications of the vanilla DeepOnet for dynamical response prediction, the initial input to the trunk network is only the vector of time steps, $\chii= \{t_i\mid i = 1, \ldots, N_t\}$.

Ultimately, the above two networks, branch and trunk, are trained to learn the solution operator. For the input function $\uu^{(q)}$ the final prediction of the DeepONet, denoted as $\mathcal{G}_\theta(\uu^{(q)})(\chii_j)$ when evaluated at any coordinate $\chii_j$, is computed as the dot product of the branch and trunk outputs plus an additional bias term $b_0$, yielding the expression in \Eqn{eq:deeponet}. 

\begin{equation} \label{eq:deeponet} 
    \mathcal{G}_{\thetaa}(\uu^{(q)})(\chii_j) = b_0 + \sum_{k=1}^p b_k(u^{(q)}(t_1), u^{(q)}(t_2)\dots,u^{(q)}(t_{N_t}))\psi_k(\chii_j), 
\end{equation}
where $b_0$ is a constant bias parameter. The coefficients, $b_k(\cdot),~k=1,\dots,p,$ are the outputs from the branch network, with external force $\uu^{(q)}(\cdot)$ measured at temporal coordinates $\{t_i\}_{i=1}^{N_t}$ as input. The bases, $\psi_k(\cdot),~k=1,\dots,p,$ are the outputs from the trunk network with the spatial-temporal coordinate $\chii_j$ as the input. Hence, the vector of trainable parameters $\thetaa$ within the DeepONet contains both weights and biases from the \textit{branch} and \textit{trunk} networks as well as the constant bias term $b_0$. 

Due to its construction, the form of the DeepONet shown in \Eqn{eq:deeponet} is capable of learning the full spatial-temporal solution operator, which is independent of the discretization used in generating the training data. The DeepONet parameters in $\thetaa$ are estimated by training over a data set $\mathcal{D}$ by solving an optimization problem for an $L_2$-loss between the DeepONet-predicted value  $G_{\thetaa}(\uu^{(q)})(\chii_j)$ and target value from the training data set $G(\uu^{(q)})(\chii_j)$, which can be expressed as

\begin{equation} \label{eq:opt}
\begin{split}
 \min\limits_{\thetaa} J(\thetaa) :&= \frac{1}{N_x\cdot N_u} \sum_{q=1}^{N_u}\sum_{j=1}^{N_x} \lvert G_{\thetaa}(\uu^{(q)})(\chii_j) - G(\uu^{(q)})(\chii_j) \rvert^2\\
 &= \frac{1}{N_x\cdot N_u} \sum_{q=1}^{N_u}\sum_{j=1}^{N_x} \left\lvert b_0 + \sum_{k=1}^p b_k(u^{(q)}(t_1),\dots,u^{(q)}(t_{N_t}))\psi_i(\chii_j) - G(\uu^{(q)})(\chii_j) \right\rvert^2
\end{split}
\end{equation} 
The solution to \Eqn{eq:opt} can be found using the Adam algorithm \cite{kingma2014adam}, for example, which is a popular variant of the stochastic gradient descent method \cite{bottou2010large,bottou2012stochastic}.  It should be noted that the measured response locations $\{\chii_j\}_{j=1}^{N_x}$ can differ from locations $\chii$ where response predictions are sought from the trained DeepONet. As a further note, while different realizations of the external forcing function $\uu^{(q)}$ can be sampled from function spaces, \eg Gaussian random fields and Chebyshev polynomials \cite{lu2021learning}, this study samples from the probability distributions of uncertain variables to instantiate new realizations of $\uu^{(q)}$. 

\subsection{Extended DeepONet (Ex-DeepONet)}\label{sec:ex-deeponet}

\begin{figure}[tb]
	\centering
	\begin{subfigure}[b]{0.48\textwidth}
		\centering
		\includegraphics[width=\linewidth]{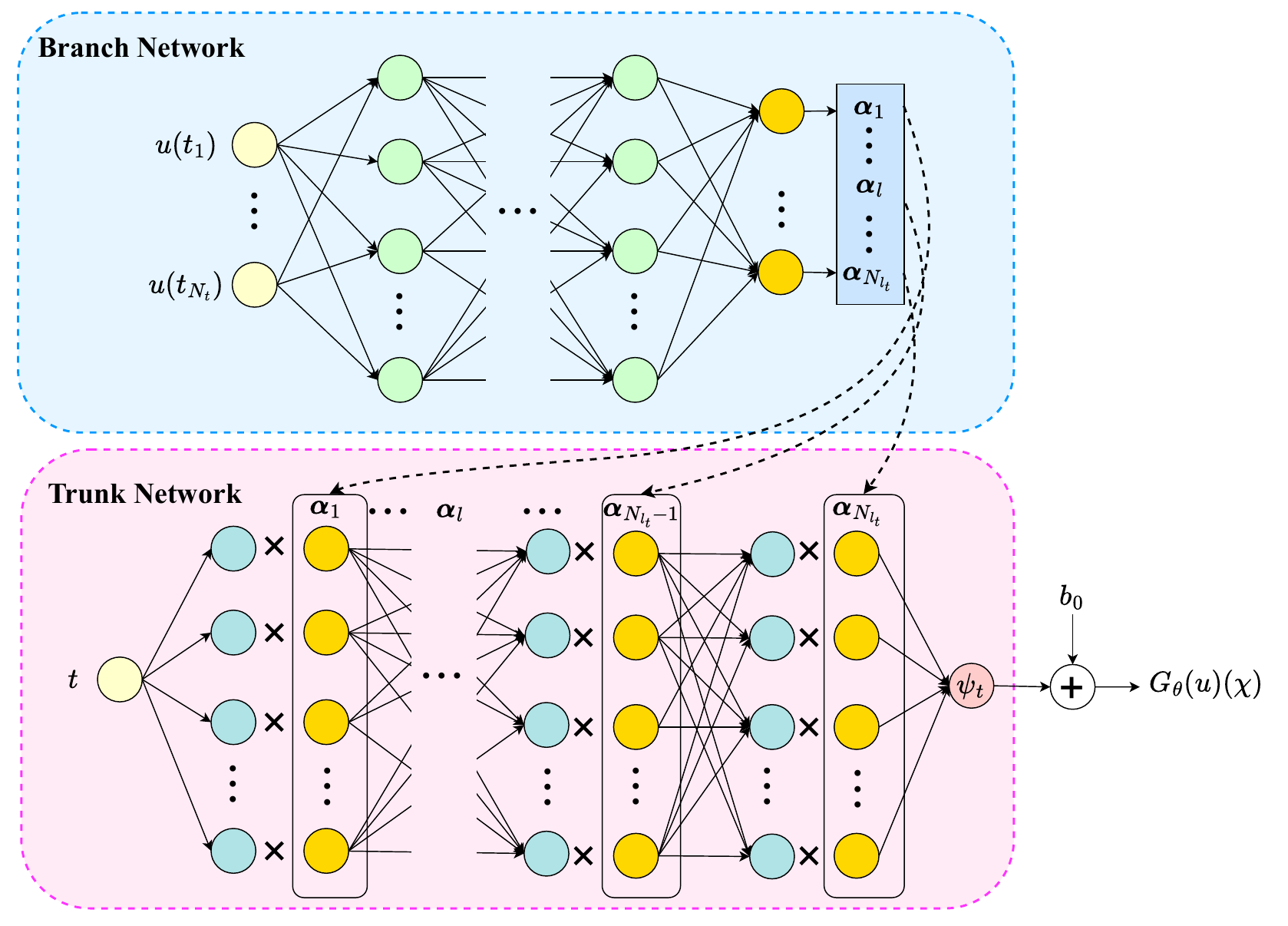}
        \caption{Ex-DeepONet for SDoF response}\label{fig:Ex-DeepONet_SDoF}
	\end{subfigure}
 \vspace{9pt}
     \begin{subfigure}[b]{0.48\textwidth}
		\centering
		\includegraphics[width=\linewidth]{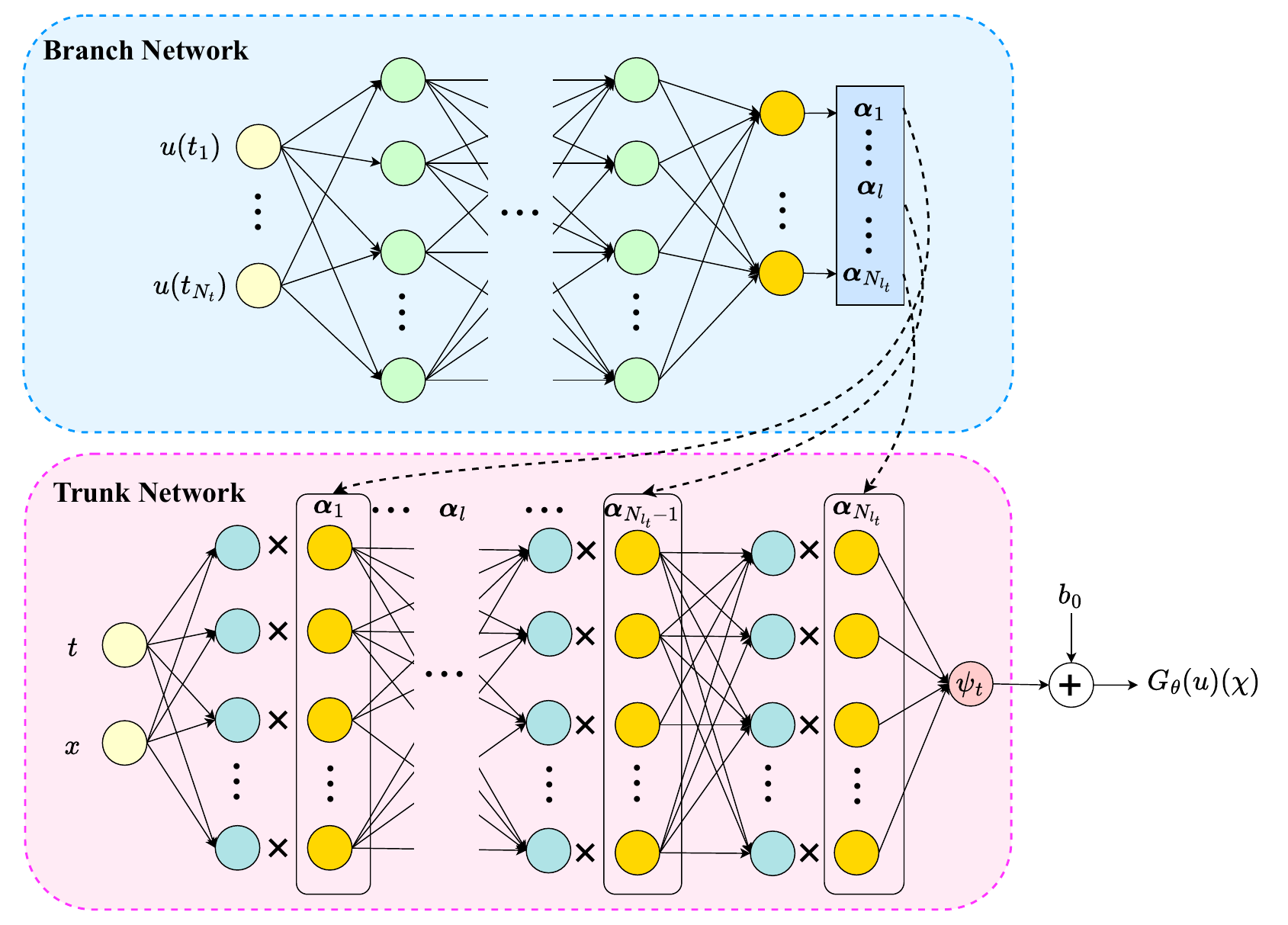}
        \caption{Ex-DeepONet for MDoF response}\label{fig:Ex-DeepONet_MDoF}
	\end{subfigure}
	\caption{Schematic illustration of an Ex-DeepONet architecture considering (a) individual DoF responses and (b) MDoF responses.}
	\label{fig:Ex_DeepONet}
\end{figure}

As stated by Li et al. \cite{li2025architectural}, the capability of vanilla DeepONet to approximate nonlinear PDE solutions is constrained by its architecture, which relies on basis expansion through dot products of branch and trunk outputs. To overcome this limitation, a generalized form of DeepONet, namely the Extended DeepONet (Ex-DeepONet) \cite{li2025architectural}, can be formulated by enabling the outputs of the branch network to interact with all the hidden layers of the trunk, not just the output layer, as depicted by the architecture in \Fig{fig:Ex_DeepONet}.

The Ex-DeepONet architectures for predicting SDoF and MDoF responses are modified according to the corresponding vanilla DeepONet architectures shown in \Fig{fig:vanilla_deeponet}. For the Ex-DeepONet, the depth of the branch network is kept the same as in \Eqn{eq:mlp_branch}, but the final layer of the branch network is modified to produce a vector $\boldsymbol{\alpha}$ with dimension $\boldsymbol{\alpha} \in \mathbb{R}^{1\times N_{l_t}N_w}$, which is given by multiplying the depth and width of trunk network. Thus, each element in $\boldsymbol{\alpha}$ corresponds to an individual neuron in the hidden layers of the trunk network.  
This vector is reshaped into a matrix $\mathbf{A}$ with column vectors $\{\boldsymbol{\alpha}_\ell \in \mathbb{R}^{N_w}\}_{\ell=1}^{N_{l_t}}$. At each hidden layer of the trunk network in the Ex-DeepONet, we multiply the linear output elementwise by $\boldsymbol{\alpha}_\ell$ before applying the activation, thereby allowing the branch network to modulate every trunk layer. 

If $\mathbf{z}_t^{\ell-1}$ is the previous trunk layer output, the Ex-DeepONet modifies \Eqn{eq:mlp_trunk} such that each trunk layer output is scaled by the branch coefficients, as formulated in \Eqn{eq:ex_trunk}
\begin{subequations}
\begin{align}
    \mathbf{A} &= [\boldsymbol{\alpha}_1, \boldsymbol{\alpha}_2, \ldots, \boldsymbol{\alpha}_\ell, \ldots, \boldsymbol{\alpha}_{N_{l_t}}] \in \mathbb{R}^{N_w \times N_{l_t}}\\
     \mathbf{z}_t^\ell&= \mathcal{R}_t^\ell[(\mathbf{W}_t^{\ell} \mathbf{z}_t^{\ell-1} + \mathbf{B}_t^{\ell}) \odot \boldsymbol{\alpha}_\ell]\\
    \mathbf{\psi}_t &= \mathbf{W}_t^{N_{l_t}+1} \mathbf{z}_t^{N_{l_t}} + \mathbf{B}_t^{N_{l_t}+1} 
\end{align}\label{eq:ex_trunk}
\end{subequations}
where $\odot$ is elementwise multiplication, and $\boldsymbol{\alpha}_\ell$ is precisely the subset of the branch-network output that corresponds to $\ell$th layer. $\mathbf{z}_t^\ell$ denotes the output features of each hidden layer $\ell = 1, 2, \ldots, N_{l_t}$, with activations adjusted by the corresponding coefficients from the branch network. 

Since the output of the branch network is incorporated into the trunk network, the final prediction of the Ex-DeepONet, $\mathcal{G}_{\thetaa}(\uu)(\chii)\in \mathbb{R}^p$, is no longer a dot product between subnetworks.  Rather, the final prediction is directly represented by the output features of the trunk network $\mathbf{\psi}_t$. 
Therefore, the dimension of the trunk output linear layer is set to $p = 1$ based on two considerations: (i) for SDoF systems, only one scalar output value is required; and (ii) for MDoF systems, due to the one-to-one correspondence between predefined spatial coordinates and their target responses, each prediction naturally corresponds to a single spatio-temporal value. 

As observed by comparing \Fig{fig:vanilla_deeponet} and \Fig{fig:Ex_DeepONet}, the Ex-DeepONet facilitates a more expressive representation of neural operators than the vanilla DeepONet through its enhanced basis formation, expanded branch interaction, and nonlinear combination. Specifically, the Ex-DeepONet retains the hierarchical characteristics previously associated with vanilla DeepONet while enabling a more adaptive feature representation through interaction between the branch network and all layers of the trunk network. This transforms the model's basis functions from a linear combination to a deeper and nonlinear interaction due to nonlinear activation functions in the trunk network. Consequently, the branch network actively contributes not only to the learning of basis coefficients but also directly influences the formation of the basis functions themselves. Thus, the Ex-DeepONet represents a more versatile and generalized neural operator architecture. 

\subsection{Full-field Ex-DeepONet}\label{sec:FExD}

\begin{figure}[tb]
    \centering
    \includegraphics[scale=0.8]{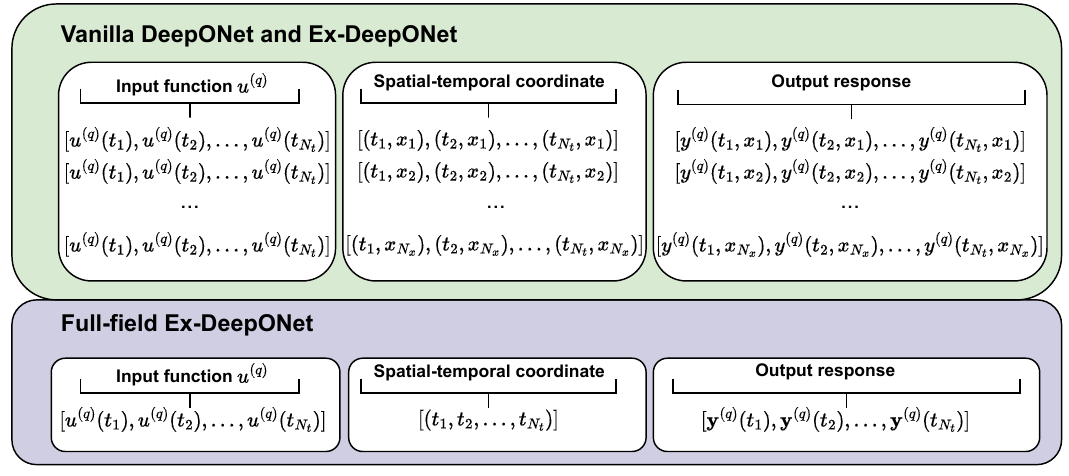}
    \caption{Data structuring for MDoF of vanilla vs. Ex-DeepONet vs. Full-field Ex-DeepONet.}\label{fig:data_pre}
\end{figure}

Despite the modifications to both the vanilla and Ex-DeepONet to create the capacity to predict multiple responses when provided with spatio-temporal coordinates, they are still limited by requiring rigid mappings between predefined spatial coordinates and individual responses. \Fig{fig:data_pre} illustrates the data structures adopted for each realization in the vanilla and Ex-DeepONets. For a complete time history of the input function $\uu^{(q)}$, the trunk network takes as input a replicated set of spatial-temporal coordinates $(t_i, x_j)$ that sweeps every spatial sensor location $x_j$ across all time instants $t_i$, while the output provides target responses $y^{(q)}(t_i, x_j)$ at the selected location $x_j$. Thus, these architectures must be separately evaluated for every individual spatial coordinate pair to recover a full field, which can limit the complex spatial-temporal dependencies inherent in dynamical systems.

\begin{figure}[tb]
    \centering
    \includegraphics[scale=.3]{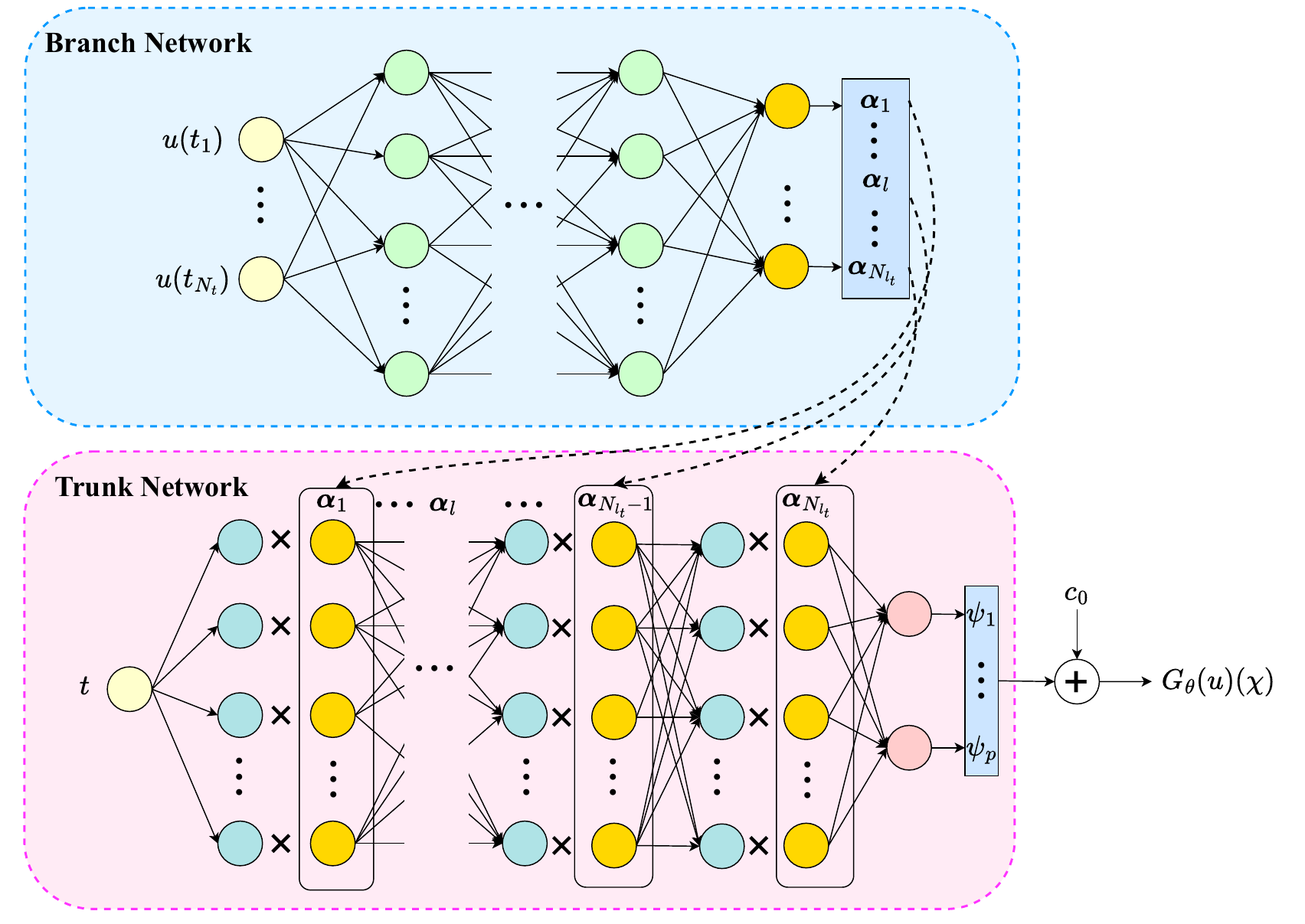}
    \caption{Full-field Ex-DeepONet.}\label{fig:FExD} 
  \end{figure}

To address this challenge, we propose the full-field Ex-DeepONet, which combines the advantages of the expressive network architecture of Ex-DeepONet with simplified dataset preprocessing that eliminates explicit spatial coordinates. This full-field Ex-DeepONet outputs the entire spatial field in a single evaluation, as illustrated in \Fig{fig:FExD}. The branch network receives only temporal measurements of input function $\uu^{(q)}$ as with the previous Ex-DeepONet; however, the trunk network is designed to only process temporal coordinates $\mathbf{t} = \{t_1, t_2, \ldots, t_{N_t}\}$ because the full-field Ex-DeepONet, via the trunk, is tasked with learning the full-field of spatial dynamical responses in a single pass.  
Thus, the dimension of the final layer, $p$, is equal to the number of spatial degrees of freedom $N_x$ being evaluated, \ie $p = N_x$, which enables simultaneous prediction of responses at all spatial locations directly. %\ie $\mathbf{\psii}_t = [\psi_1, \psi_2, \ldots, \psi_{N_x}]$. 
Adopting this approach efficiently yields vector-valued predictions $G_{\thetaa}(\uu^{(q)})(t_i) \approx \mathbf{y}^{(q)} \in \mathbb{R}^{N_x}$. Note that this architecture reduces to the Ex-DeepONet for a SDoF response if the output dimension $p$ is set to 1.

The full-field Ex-DeepONet excels in spatio-temporal surrogate modeling by addressing limitations of the aforementioned vanilla DeepONet and covnentional Ex-DeepONet while offering architectural flexibility that eliminates complex data preprocessing requirements for each spatial location. These advantages make it an efficient and expressive approach for modeling high-dimensional dynamical systems, as demonstrated by the application presented in the following section.

\section{Application to Structural Dynamics}

The capabilities of the proposed full-field Ex-DeepONet, as well as those of the vanilla DeepONet and conventional Ex-DeepONet, are interrograted through a structural dynamics application. 
A cable-stayed bridge serves as a “testbed” dynamical system for evaluating these approaches. %validating three aforementioned approaches. 
Numerous realizations of the bridge response are generated by subjecting the bridge to stochastic ground motions. These realizations form the basis of the training data for the proposed approaches. \Sec{sec:strdyn} provides some background on connecting the governing equations of structural dynamics to operator learning.  Sections~\ref{sec:bridge} and \ref{sec:input} provide details on the bridge model and the input excitation, respectively.

\subsection{Structural Dynamics}\label{sec:strdyn}

For MDOF dynamical systems subjected to a ground motion, the equation of motion is given by \Eqn{eq:beamEOM}

\begin{equation} \label{eq:beamEOM}
    \mathbf{M} \ddot{\mathbf{y}}(t) + \mathbf{C} \dot{\mathbf{y}}(t) + \mathbf{K} \mathbf{y}(t) = \mathbf{f}(t) = -\mathbf{M} \mathbf{r} \ddot{u}_g(t)
\end{equation}
where $\mathbf{M}$, $\mathbf{C}$, and $\mathbf{K}$ are the mass, damping, and stiffness matrices of the system, respectively; $\mathbf{y}$, $\dot{\mathbf{y}}$, $\ddot{\mathbf{y}}$ are the generalized vectors of displacement, velocity, and acceleration, respectively;  $\ddot{u}_g$ is the ground acceleration in the longitudinal direction; and $\mathbf{r}$ is the influence vector for the ground acceleration (i.e., it is a vector that consists of ones corresponding to active horizontal longitudinal displacement DOFs and zeros elsewhere).

For linear systems, the response $\mathbf{y}(t)$ in \Eqn{eq:beamEOM} to an arbitrary input $\ddot{u}_{g}(t)$ can be derived by a convolution integral shown in \Eqn{eq:resp}. This follows from the superposition principle, whereby the total response of the system is the sum of responses to infinitesimal impulses $\ddot{u}_{g}(\tau)$
 
\begin{equation}\label{eq:resp}
\bm{y}(t)\equiv\int_{0}^{t}\ddot{u}_{g}(\tau)\bm{h}(t-\tau)d\tau
\end{equation}
where \(\bm{h}(t-\tau)\) is the impulse response function, namely the Green's function, that represents the system’s response at time $t$ due to a unit impulse at time $\tau$. 

It can be readily observed that the structural dynamics relation shown in \Eqn{eq:resp} conforms to the expression in \Eqn{eq:gov}. 
The solution to this governing equation, the structural response, can be expressed as an operator $y(\chii) = \mathcal{G}(\eta)(\chii)$, \ie the solution to \Eqn{eq:resp}, where $\eta = \ddot{u}_{g}$. Thus, the solution operator represents a mapping from ground acceleration to the structural response across the full range of degrees of freedom: $\mathcal{G}:\ddot{u}_g(t)\longrightarrow \mathbf{y}(t)$. For initial value problems, such as those encountered in structural dynamics, initial conditions $y(\chii_0) = y_0$ must also be specified. The objective of this study is to evaluate the capacity of the proposed full-field Ex-DeepONet, as well as the vanilla DeepONet and Ex-DeepONet, to approximate the full-field solution operator and therefore predict the spatial-temporal response of the bridge structure.  %by examining the potential capacity of the Full-field Ex-DeepONet against the vanilla and Ex-DeepONet.

\subsection{Bridge Model} \label{sec:bridge}

\begin{figure}[tb]
    \centering
    \begin{subfigure}{\textwidth}
        \centering
        \includegraphics[width=0.8\linewidth]{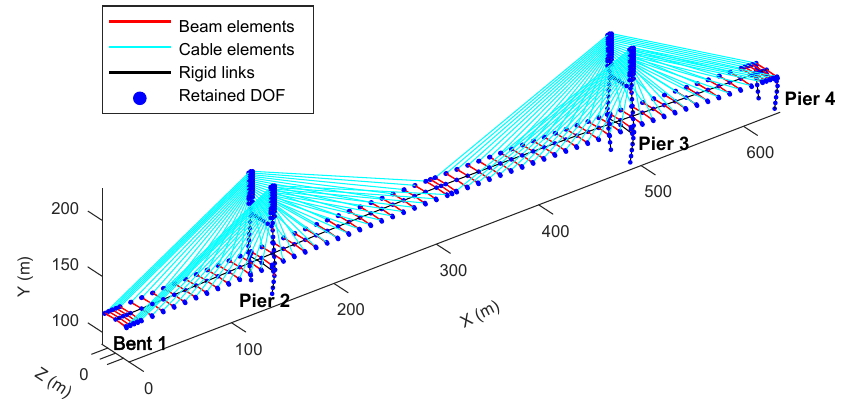}
        \caption{Finite element model of the bridge (dimensions in meters).}\label{fig:fe_model} 
    \end{subfigure}
    \hfill
    \begin{subfigure}{\textwidth}
        \centering
        \includegraphics[width=0.8\linewidth]{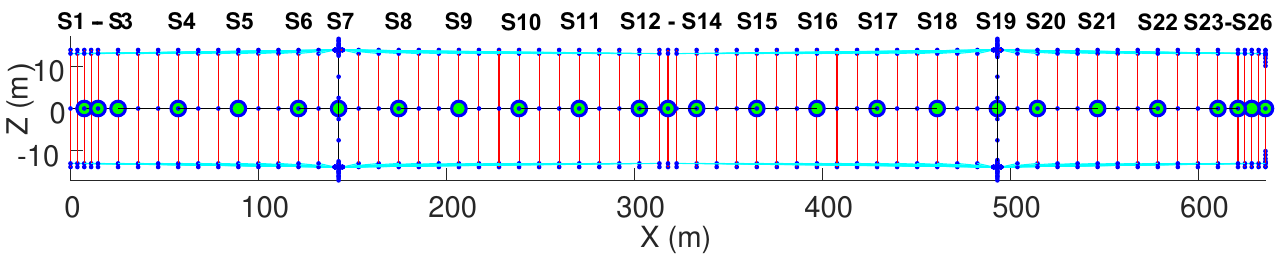}
        \caption{Top view of the bridge deck from the finite element model showing the locations of the sensor positions} \label{fig:fe_top}
        \end{subfigure}
        \caption{Finite element model of the bridge structure showing (a) the full model and (b) top view of bridge deck with sensor positions.}
	
\end{figure}

The cable-stayed bridge ``testbed'' is a benchmark finite element (FE) model of the Bill Emerson Memorial Bridge \cite{dyke2003phase}, which was built in 2003 to span the Mississippi river between Cape Girardeau, Missouri, and East Cape Girardeau, Illinois. 
The FE model, which is shown in Figure \ref{fig:fe_model}, consists of 579 nodes, 128 cable elements, 162 beam elements, 420 rigid links, and 134
nodal masses. It has been subsequently utilized in extensive studies within the structural control and health monitoring community \cite{FISCO2011275,Fujino01082013,de2017efficient}. The full 3474 degree-of-freedom (DOF) model can be reduced down to 909 DOFs after imposing boundary conditions and removing slave DOFs. A static condensation can also be applied to further eliminate DOFs with small contributions to the global response, as described in the original benchmark paper \cite{dyke2003phase}. Ultimately, the final FE model contains 419 DOFs, which is the configuration used for the following application and investigation. Response measurements at each node, such as relative displacement, velocity, and absolute acceleration in the longitudinal (X), vertical (Y) and transverse (Z) directions,
can be obtained through numerical simulations. 

Phase II of the original benchmark study \cite{caicedo2003phase} identified 27 sensing locations along the longitudinal axis of the bridge deck that included approximately every third node along the bridge deck centerline. These 27 positions were intentionally selected to include the side-span, mid-deck, piers, and approximately 1/3 and 1/6 points. This study assumes that responses in the vertical direction can be recovered but only considers 26 of those original 27 sensing locations, which are highlighted in \Fig{fig:fe_model} and shown with labels in \Fig{fig:fe_top}.  The motion at bent 1 is excluded from the selected locations because its boundary conditions constrain its movement, only allowing longitudinal (X) displacement and rotations about the Y and Z axes. 
The vertical accelerations at the 26 sensor locations will constitute the target responses for the various DeepONet models in the subsequent analysis. \Tab{tab:prim_freq} lists the first few modes and natural frequencies identified using the vertical acceleration data. 

\begin{table}[tb]
    \centering
    \caption{Primary identified natural frequencies of the benchmark cable-stayed bridge}\label{tab:prim_freq}
    \scriptsize
    \begin{tabular}{c|c|c}
        \toprule
        Mode No. & Description & Frequency (Hz)\\
        \midrule
         1 & First vertical mode & 0.2899 \\
         2 & Second vertical mode & 0.3699 \\
         3 & Third vertical mode & 0.5812 \\
         4 & First lateral-torsional mode & 0.6490 \\ 
        \bottomrule
    \end{tabular}
\end{table}

\subsection{Stochastic Ground Motion Excitation} \label{sec:input}

The bridge model is exctied by sujecting it to stochastic ground motions.  The
ground motions are simulated by multiplying stationary excitation with an envelope function, where the spectral density of the stationary excitation is modeled by the Kanai–Tajimi (K-T) spectrum \cite{guo2016system}. Assuming that ground acceleration \(\ddot{u}_g\) follows a zero-mean Gaussian distribution, the K-T spectrum can be described as in \Eqn{eq:Kanai-Tajimi}
\begin{subequations}\label{eq:Kanai-Tajimi}
    \begin{align}
        S(\omega) &= S_0 \frac{\omega_g^4 + (2\zeta_g \omega_g \omega)^2}{(\omega_g^2 - \omega^2)^2 + (2\zeta_g \omega_g \omega)^2}\\
        S_0 &= \sigma_g^2 \frac{2\zeta_g}{\pi \omega_g (4\zeta_g^2 + 1)}
    \end{align}
\end{subequations}
where $\omega_g$ is the dominant frequency of excitation, $\zeta_g$ is the bandwidth of the excitation, and $\sigma_g$ is the standard deviation of the excitation (assuming a two-sided spectrum). In this study, the parameters for the Kanai-Tajimi spectrum are uniformly distributed according to $\sigma_g \sim U$ (0.8, 1.0) and $\zeta_g \sim U$ (0.2, 0.4), while $\omega_g$ is kept as $10\pi$ \cite{guo2016system}.

The time-varying intensity of ground motion is controlled by the envelope function, as shown in \Eqn{eq:envelope}

\begin{equation}\label{eq:envelope}
    E(t) = a \left( \frac{t}{t_n} \right)^b \exp\left(-c \cdot \frac{t}{t_n}\right)
\end{equation}
where $b = -\frac{\epsilon_t \cdot \ln \eta}{1 + \epsilon_t \cdot (\ln \eta - 1)}, c = \frac{b}{\epsilon_t}, a = \left( \frac{e}{\epsilon_t} \right)^b$, $t_n$ is the duration of strong ground motion, $\epsilon_t$ is the normalized duration time when ground motion achieves peak, and $\eta$ is the fraction of the peak amplitude at 90 percent of the duration. The parameters of the time domain envelope function are chosen as $\eta = 0.05$ and $t_n = 2$~s, while $\epsilon_t$ follows the uniform distribution with $\epsilon_t \sim U$ (0.1, 0.2) \cite{guo2016system}.

The complete ground acceleration time history is generated by employing a linear superposition of a large number of harmonic processes at varying frequencies, as expressed in 
\begin{equation}\label{eq:ground_acc}
\ddot{u}_g(t_j) = \sum_{n=1}^{N} E(t_j) A_n \cos(\omega_n t_j + \theta_n), \quad j = 0, 1, \dots, N-1,
\end{equation}
where $A_n = \sqrt{2S(\omega_n)\Delta \omega}$, in which $S(\omega_n)$ is the spectrum parameterized by the frequency $\omega_n$, %of excitation, 
and $\theta_n$ are randomly generated phase angles uniformly distributed between $0$ and $2\pi$.

\section{Implementation Details for MDoF responses}

\begin{figure}
	\centering
	\includegraphics[width=0.5\linewidth]{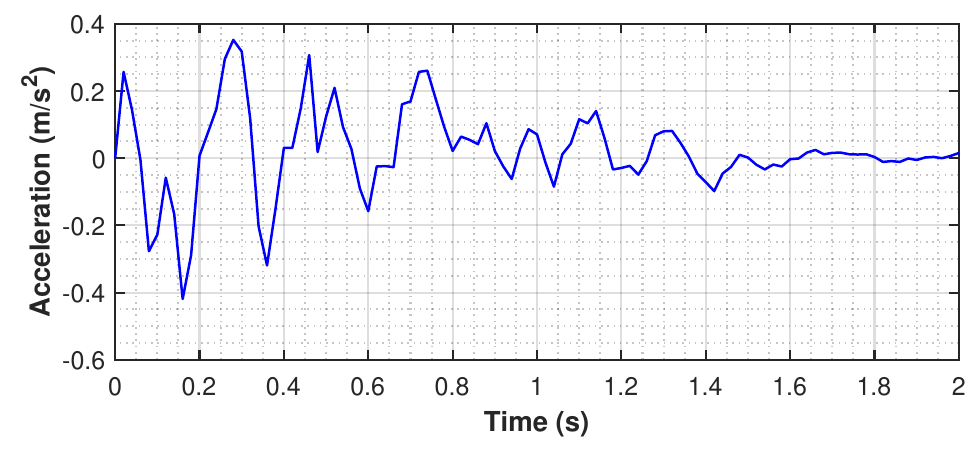}
	\caption{Sample ground motion realization.} \label{fig:sample_ug}
\end{figure}

We generate $3\times 10^4$ realizations of ground accelerations by sampling from the distributions of the uncertain parameters that parameterize the K-T spectrum and evelope function.  The ground motions are discretized at a sampling frequency of 50 Hz over a duration of 2 seconds. The collection of ground motions serve as inputs to the branch network. A sample realization is shown in \Fig{fig:sample_ug}. 
The corresponding structural acceleration responses at the 26 specified sensor locations are collected as the target spatial-temporal outputs for the DeepONets. Further, it is assumed that the structure is starting from a quiescent condition; thus, the initial conditions are identically set to zero.

The acceleration response realizations are split using mean-based stratified sampling \cite{aviles2024stratified} to ensure representative distributions into a ratio of 60:20:20 for training, validation, and testing, respectively. For each complete model realization, a single mean value is calculated by averaging across all time steps and 26 spatial positions. These mean values are then grouped into quantile-based bins. This data splitting strategy enables evaluation of model performance on complete, physically reasonable cases across all datasets. The training data is normalized using a min-max scaling with a range of $[-1,1]$; the validation and testing data subsets are subsequently normalized based on the scaling used during training.

The acceleration responses, along with the spatial-temporal coordinates and input functions, \ie $\ddot{u}_g(t)$, form the complete data sets used for training the vanilla DeepONet, Ex-DeepONet, and full-field Ex-DeepONet; however, the data structuring differs based on network architecture. Specifically, the trunk of the vanilla and Ex-DeepONet takes as input spatio-temporal coordinates $(t_i, x_j)$. Across time instances and spatial locations, this yields $(N_u \times N_t \times N_x)$ samples, where each sample forms a tuple with a shape of $\{[\mathbf{u}^{(q)} \in \mathbb{R}^{N_t}], [(t_i, x_j) \in \mathbb{R}^2], [y^{(q)}(t_i, x_j) \in \mathbb{R}]\}$. In contrast, the trunk of the full-field Ex-DeepONet processes only temporal coordinates. Moreover, the full-field Ex-DeepONet outputs vector-based responses. Across time instances, this yields $(N_u \times N_t)$ samples, where each sample forms a tuple with a shape of $\{[\mathbf{u}^{(q)} \in \mathbb{R}^{N_t}], [t_i \in \mathbb{R}], [\mathbf{y}^{(q)}(t_i) \in \mathbb{R}^{N_x}]\}$. This structured approach distinguishes these DeepONet methodologies from datasets employed in recurrent neural networks (RNNs) \cite{ryan2021machine} and convolutional neural networks (CNNs) \cite{Goodfellow-et-al-2016}, which lack trunk inputs and typically follow a more conventional feature-target format. It is worth mentioning that when training for SDoF response prediction, the trunk network for every architecture only always takes time instances as input since there is only a single spatial location.

\subsection{DeepONets Training}

This section discusses the training procedure for the three models, \ie (i) the vanilla DeepONet (referred to as ``VD''), (ii) the Ex-DeepONet incorporating spatiotemporal coordinates (referred to as ``ExD''), and (iii) the full-field Ex-DeepONet (referred to as ``FExD''). 

\subsubsection{Model architectures and hyperparameters} \label{sec:train}

The architectures of each DeepONet are designed through a series of parametric experiments to optimize hyperparameters. These experiments explored various combinations of network depths and widths, specifically $N_{l_t}$ for trunk networks and $N_{l_b}$ for branch networks, as well as the number of neurons per layer. For fair comparison, we maintain consistent network dimensions and training hyperparameters across all models, although the branch and trunk networks can be independently configured with different depths and widths. Starting with a baseline architecture of 4 layers with 50 neurons each, which yielded relatively high training and validation loss, we increased model complexity by incrementally adjusting both the number of layers and neurons per layer to improve performance. Key findings indicate that a dropout rate of zero outperforms non-zero rates, e.g., 0.01 to 0.2. Similarly, increasing the depth to 8 or 10 layers results in poorer performance, underscoring the challenges of training deeper networks. In contrast, wider architectures enhance learning capacity but introduce overfitting risks. Upon completion of the parametric study, the suggested architecture for three DeepONets features branch and trunk networks with four layers, with each hidden layer containing 300 neurons.  Note that the output dimensions are unique to each DeepONet due to inherent differences in architecture, as detailed in \Tab{tab:DeepONet_prm}. 

\begin{table}[tb]
    \centering
    \scriptsize
    \caption{DeepONet size for each case, unless otherwise stated.}
    \label{tab:DeepONet_prm}
    \begin{tabular}{l|cccc ccc cc}
      \toprule
      \multirow{2}{*}{Location} & \multirow{2}{*}{Network type} 
          & \multicolumn{3}{c}{Branch} & \multicolumn{3}{c}{Trunk} & \multirow{2}{*}{Epoch} & \multirow{2}{*}{Time $(h)$} \\
      \cmidrule(lr){3-5}\cmidrule(lr){6-8}
          & & Depth & Width & Output dimension & Depth & Width & Output dimension & & \\
      \midrule
      \multirow{3}{*}{Multiple} 
          & VD           & 4 & 300  & 10  & 4 & 300  & 10 & 520 & 13.45\\
          & ExD          & 4 & 300  & 1200  & 4 & 300  & 1 & 783 & 20.85 \\
          & FExD         & 4 & 300  & 1200 & 4 & 300 &26 & 988 & 0.92 \\
      \bottomrule
    \end{tabular}
  \end{table}

The mean squared error ($L_2$-norm) loss function is used as the optimization objective during training, as formulated in \Eqn{eq:opt}. To effectively mitigate overfitting and enhance generalization, we employ a three-pronged optimization strategy: (i) the AdamW optimizer with weight decay \cite{hutter} that uses a nominal $L_2$-norm regularization coefficient of $10^{-8}$; (ii) a cyclical learning rate scheduler that dynamically varies the learning rate between $10^{-4}$ and $10^{-3}$ throughout training, allowing the model to descend into finer features of the loss landscape during low learning rate phases while maintaining the ability to escape local minima during high learning rate phases; and (iii) an early stopping mechanism, triggered if improvements in validation loss remain below 1\% for 50 consecutive epochs. While training for additional epochs may not benefit models prone to overfitting, increasing the dataset size is highly effective in improving performance across both training and validation sets.

\subsubsection{Model training}

\begin{figure}[tb]
    \centering
    \begin{subfigure}{0.32\textwidth}
        \centering
        \includegraphics[width=\linewidth]{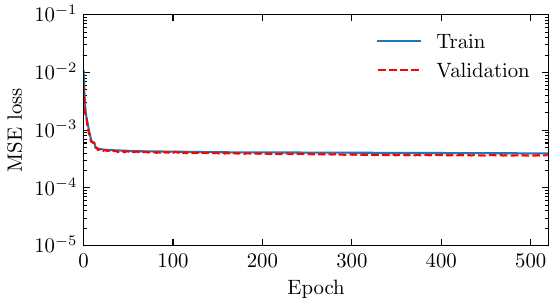}
        \caption{VD model}
    \end{subfigure}
    \hfill
    \begin{subfigure}{0.32\textwidth}
        \centering
        \includegraphics[width=\linewidth]{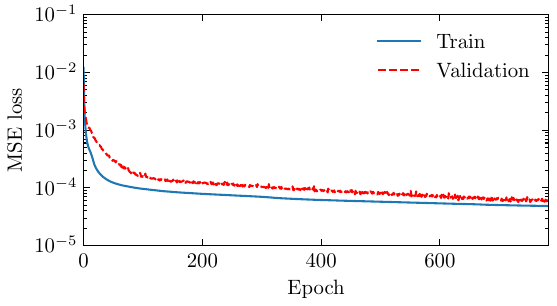}
        \caption{ExD model}
    \end{subfigure}
    \hfill
    \begin{subfigure}{0.32\textwidth}
        \centering
        \includegraphics[width=\linewidth]{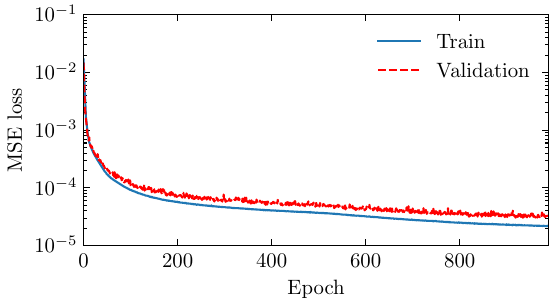}
        \caption{FExD model}
    \end{subfigure}
    \caption{Training and validation mean squared error (MSE) losses for each of the considered DeepONet models.}
    \label{fig:loss}
\end{figure}

The three models were implemented using the Python/PyTorch framework, and training and testing were conducted on an Nvidia RTX A2000 GPU card with a batch size of 256. \Fig{fig:loss} illustrates the convergence characteristics of the VD, ExD, and FExD models during the training process. Following the initial optimization phase over the first few epochs, the VD model exhibits limited capacity to capture complex patterns, resulting in a performance plateau. In contrast, the FExD model demonstrates superior convergence efficiency and achieves lower overall error metrics relative to both VD and ExD architectures. All three models exhibit excellent generalization capabilities, as evidenced by the negligible discrepancy between their respective training and validation loss trajectories.

The training time of the three DeepONet architectures is quantitatively compared in the last column of \Tab{tab:DeepONet_prm}. Note that the models require different numbers of epochs to converge for the reasons described in Section~\ref{sec:train}, as shown in the second to last column in \Tab{tab:DeepONet_prm}. The FExD model demonstrates superior convergence efficiency, requiring only 0.92 hours to complete training, which represents a significant reduction compared to the ExD model with 20.85 hours and the VD model with 13.45 hours. The FExD architecture's ability to process multiple spatial coordinates simultaneously in a single forward pass leads to a substantial acceleration in training that is approximately 22.66 times faster than the ExD and 14.62 times faster than the VD. Despite needing more epochs to converge than ExD and VD, the per-epoch computational cost for FExD is dramatically lower. This efficiency stems from its functionally enriched architecture that eliminates the need for separate evaluations at each spatial coordinate, which is particularly advantageous when modeling complex dynamical systems with multiple DoFs. The more straightforward dataset formation and more expressive architecture allow the FExD to maintain superior training and validation error metrics while significantly reducing the computational burden associated with training traditional DeepONets for surrogate modeling applications.

\section{Results and Discussion}

This section presents and compares the performance of the three DeepONet approaches when applied to testing data, \ie realizations not seen during training or validation.% of VD, ExD, and FExD in terms of both time- and frequency-domain response prediction.

\subsection{Prediction error criteria}

To quantitatively evaluate the performance of DeepONet-based surrogates, four evaluation criteria are computed. The first three criteria are mean absolute error (MAE), root mean squared error (RMSE), and relative root mean squared error (RRMSE), which measure the difference between the target (ground truth) and surrogate predictions. The fourth evaluation criterion is the coefficient of determination (${R}^2$), which indicates how well the model fits the data and expresses this as a percentage, with zero indicating a very poor fit and 100\% indicating a perfect fit. These error metrics are defined in \Eqn{eq:error_metric}

\begin{subequations} \label{eq:error_metric}
\begin{align}
    \mathrm{MAE} &= \frac{1}{N_t} \sum_{i=1}^{N_t}\left|\widehat{y}_{i}- y_{i}\right| \\
    \mathrm{RMSE} &= \sqrt{\frac{1}{N_t} \sum_{i=1}^{N_t}\left(\widehat{y}_{i}- y_{i}\right)^2} \\
    \mathrm{RRMSE} &= \sqrt{\frac{\sum_{i=1}^{N_t}\left(\widehat{y}_{i}- y_{i}\right)^2}{\sum_{i=1}^{N_t} {y_{i}}^2}} \times 100 \%\\
    R^2 &= \left\{1-\frac{\sum_{i=1}^{N_t}\left(\widehat{y}_{i}-y_{i}\right)^2}{\sum_{i=1}^{N_t}\left(\widehat{y}_{i}-\overline{y}\right)^2} \right\} \times 100\%
\end{align}
\end{subequations}
where $N_t$ is the number of data points (time steps) and $\widehat{y}_{i}$ and $y_{i}$ denote the predicted and targeted response at the $i$th time step, respectively, while $\overline{y}$ represents the time-average of the response prediction.

\subsection{Predictive performance comparison}

\begin{figure}[tb]
    \centering
    \begin{subfigure}[b]{0.3\textwidth}
        \centering
        \includegraphics[width=\linewidth]{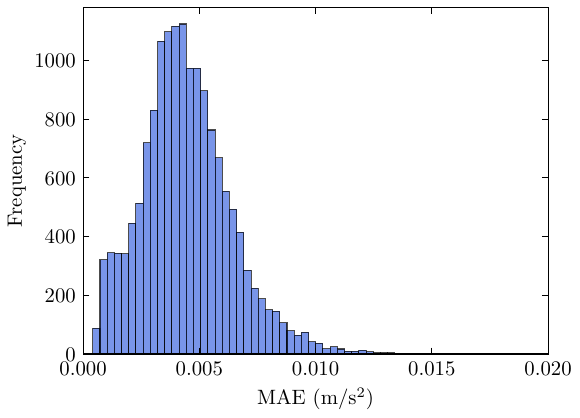}
        \caption{Testing MAE for VD}
        \label{fig:vd_mae_hist}
    \end{subfigure}
    \hspace{0.02\textwidth}
    \begin{subfigure}[b]{0.3\textwidth}
        \centering
        \includegraphics[width=\linewidth]{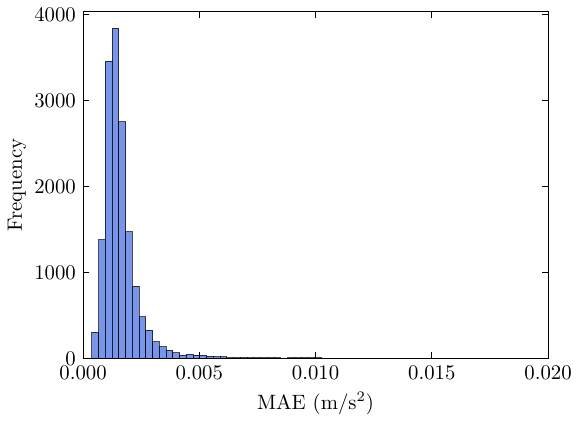}
        \caption{Testing MAE for Ex-DeepONet}
        \label{fig:exd_mae_hist}
    \end{subfigure}
    \hspace{0.02\textwidth}
    \begin{subfigure}[b]{0.3\textwidth}
        \centering
        \includegraphics[width=\linewidth]{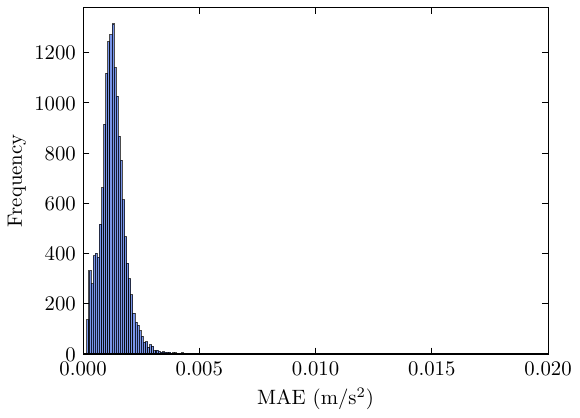}
        \caption{Testing MAE for FExD}
        \label{fig:fexd_2s_mae_hist}
    \end{subfigure}
    \caption{Comparison of testing MAE error for the chosen spatial-temporal DeepONets.}
    \label{fig:mae_hist}
\end{figure}

\begin{figure}[tb]
    \centering
    \begin{subfigure}[b]{0.3\textwidth}
        \centering
        \includegraphics[width=\linewidth]{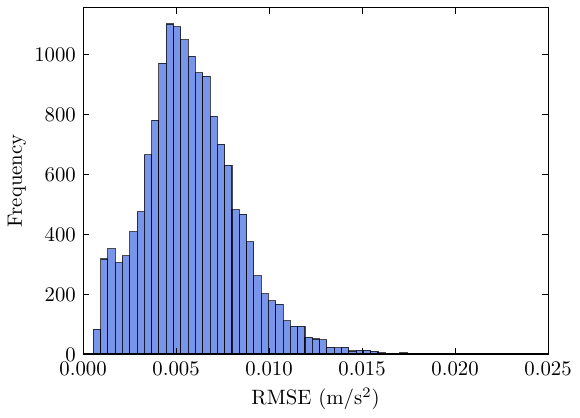}
        \caption{Testing RMSE for VD}
        \label{fig:vd_rmse_hist}
    \end{subfigure}
    \hspace{0.02\textwidth}
    \begin{subfigure}[b]{0.3\textwidth}
        \centering
        \includegraphics[width=\linewidth]{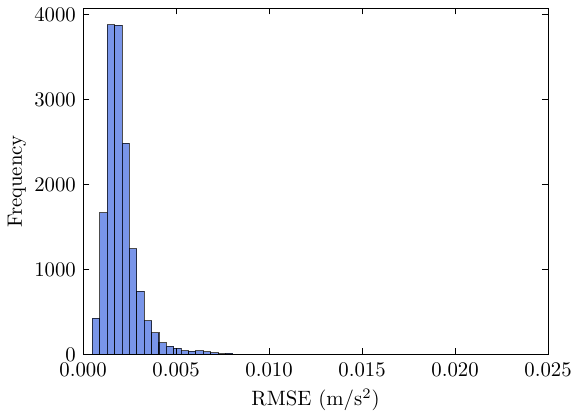}
        \caption{Testing RMSE for Ex-DeepONet}
        \label{fig:exd_rmse_hist}
    \end{subfigure}
    \hspace{0.02\textwidth}
    \begin{subfigure}[b]{0.3\textwidth}
        \centering
        \includegraphics[width=\linewidth]{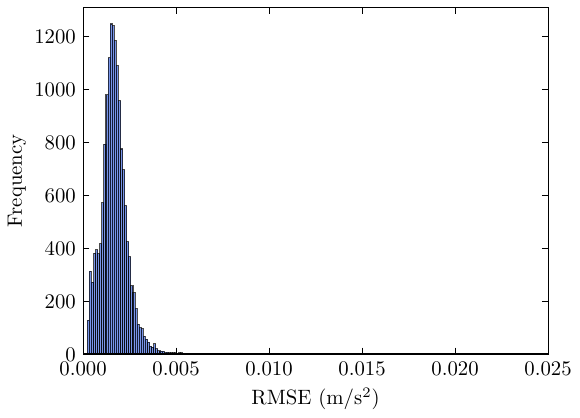}
        \caption{Testing RMSE for FExD}
        \label{fig:fexd_rmse_hist}
    \end{subfigure}
    \caption{Comparison of testing RMSE error for the chosen spatial-temporal DeepONets.}
    \label{fig:rmse_hist}
\end{figure}

\begin{figure}[tb]
    \centering
    \begin{subfigure}[b]{0.3\textwidth}
        \centering
        \includegraphics[width=\linewidth]{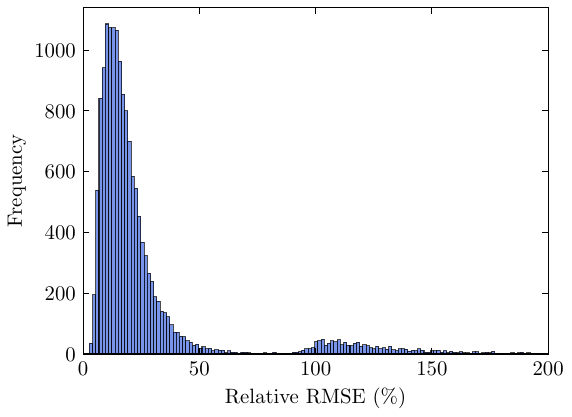}
        \caption{Testing RRMSE for VD}
        \label{fig:vd_rrmse_hist}
    \end{subfigure}
    \hspace{0.02\textwidth}
    \begin{subfigure}[b]{0.3\textwidth}
        \centering
        \includegraphics[width=\linewidth]{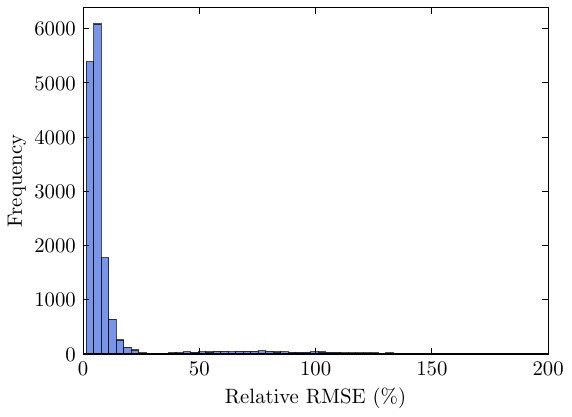}
        \caption{Testing RRMSE for Ex-DeepONet}
        \label{fig:exd_rrmse_hist}
    \end{subfigure}
    \hspace{0.02\textwidth}
        \begin{subfigure}[b]{0.3\textwidth}
        \centering
        \includegraphics[width=\linewidth]{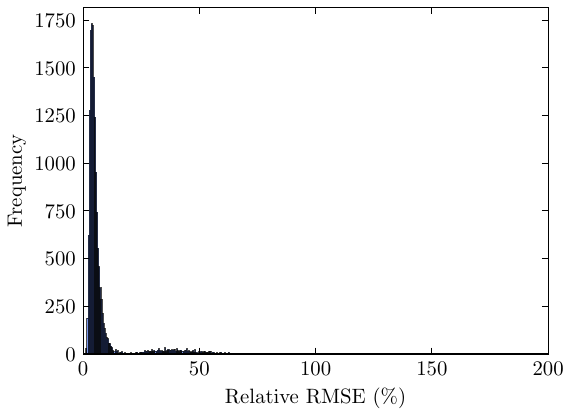}
        \caption{Testing RRMSE for FExD}
        \label{fig:fexd_2s_rrmse_hist}
    \end{subfigure}
    \caption{Comparison of testing RRMSE error for the chosen spatial-temporal DeepONets.}
    \label{fig:rrmse_hist}
\end{figure}

The various error metrics for predicting the vertical acceleration responses at the 26 sensor locations are evaluated over 600 test realizations, \ie over 15,600 individual dynamical response predictions.  Histograms of the MAE, RMSE, and RRMSE metrics are presented in \Fig{fig:mae_hist}, \Fig{fig:rmse_hist}, and \Fig{fig:rrmse_hist}, respectively. 

As observed from \Fig{fig:mae_hist} and \Fig{fig:rmse_hist}, among the three considered models, the VD exhibits notably broader error distributions for both the MAE and RMSE metrics. These distributions for the VD suggest inconsistent prediction quality as well as  limitations in the model's capacity to capture the potentially complex spatial-temporal dynamics inherent in structural responses. In contrast, both the EXD and FExD display markedly narrower error distributions, with predominant clustering of errors below $0.002 $~m/s$^2$ for MAE and $0.003 $~m/s$^2$ for RMSE. This concentration of lower error values reveals a significant enhancement in prediction stability and accuracy. Particularly noteworthy is the FExD model's error histograms, which display the sharpest peaks at the lowest error values and possess minimal high-error outliers, highlighting its precision and substantiating the superior performance previously observed during training and validation. %in \Tab{tab:error_metrics}.

Furthermore, the RRMSE histograms in \Fig{fig:rrmse_hist} display distinct patterns of error distribution across the three models. The VD presents a clear bi-modal distribution for the RRMSE, with a notable frequency of higher RRMSE values, \ie RRMSE values greater than 100\%. In contrast, the RRMSE distributions for both the ExD and especially the FExD are predominantly concentrated around lower values, with FExD achieving mostly values below 20\%.  However, both the ExD and FExd also display some bi-modality as well, albeit at lower error values than that for the VD. 

These second modes in the RRMSE distributions correspond to predictions near structural discontinuities, such as the piers and towers, where response amplitudes are inherently small. These larger values arise because the normalization in the RRMSE magnifies even minor absolute errors when the denominator, \ie target response magnitude, is small.  This explains why the RMSE distributions do not exhibit the same bi-modality. Nonetheless, the FExD still demonstrates remarkable improvement in these challenging regions, as it shifts the extreme error values to magnitudes less than half of those observed in the VD model. The substantial performance enhancement of the ExD and especially the FExD over the VD indicates their superior capacity to accurately capture both global trends and local dynamic features of structural responses, even in regions with naturally small amplitudes.

\Tab{tab:error_metrics} summarizes the average error metrics for each of the three DeepONet models.  Given that the error histograms shown in \Fig{fig:mae_hist}, \Fig{fig:rmse_hist}, and \Fig{fig:rrmse_hist} do not conform to Gaussian distributions, median error values are reported. However, the mean value and standard deviation can still be informative for comparative purposes and are, thus, featured in \Tab{tab:error_metrics} as well. 
As expected, the VD model possesses the largest average and median values across each error metric, often by a wide margin. The VD also boasts the largest standard deviation, highlighting its deficiencies in both accuracy and stability as compared to the ExD and FExD.  Another interesting observation from \Tab{tab:error_metrics} is that while the performance gains in the FExD appear modest for MAE and RMSE compared to the ExD --- around a 19\% reduction in a relative sense --- the RRMSE more clearly shows the superiority of the FExD. The FExD reduces the mean RRMSE by nearly 40\% (in a relative sense) compared to the ExD and reduces the standard deviation of the RRMSE by over 50\%. Even on its own, the FExD provides impressive results by returning an average RRMSE of less than 8\% considering the distribution includes over 15,000 individual responses.

The other metric included in \Tab{tab:error_metrics} is $R^2$, which is given in the last column.  This metric also highlights the strong performance gains achieved with the FExD as its $R^2$ values are substantially larger than those for both the VD and the ExD.  Further, the reduction in the variation in $R^2$ is also notable given that the conventional ExD performs the poorest in this regard.  Thus, the FExD manages to not only deliver a better average fit of the data, it also dramatically reduces the variation in the fit, even when compared to the ExD.  Taken all together, these various error metrics demonstrate that the FExD model can simultaneously predict the entire spatial fields with superior performance, even without explicitly predefining point-wise data structures. 

\begin{table}[tb]
    \centering
    \scriptsize
    \caption{Comparison of average error metrics across different models}
    \label{tab:error_metrics}
    \begin{tabular}{l|ccc|ccc|ccc|ccc}
    \toprule
    \multirow{2}{*}{Model} 
    & \multicolumn{3}{c|}{MAE (m/s$^2$)} 
    & \multicolumn{3}{c|}{RMSE (m/s$^2$)} 
    & \multicolumn{3}{c|}{RRMSE (\%)} 
    & \multicolumn{3}{c}{$R^2$ (\%)} \\
    & Mean & Median & Std 
    & Mean & Median & Std 
    & Mean & Median & Std 
    & Mean & Median & Std \\
    \midrule
    VD & $0.0045$ & $0.0043$ & $0.0013$ & $0.0058$ & $0.0056$ & $0.0017$ & $26.20$ & $24.37$ & $9.10$ & $83.33$ & $85.76$ & $8.88$\\
    ExD & $0.0016$ & $0.0014$ & $0.0008$ & $ 0.0021$ & $0.0019$ & $0.0010$ & $12.61$ & $11.59$ & $5.04$ & $91.80$ & $93.87$ & $10.97$\\
    FExD & $0.0013$ & $0.0012$ & $0.0004$ & $0.0017$ & $0.0016$ & $0.0005$ & $7.77$ & $7.35$ & $2.42$ & $98.29$ & $98.52$ & $1.00$\\
    \bottomrule
    \end{tabular}
\end{table}

\subsubsection{Evaluating error metrics at each position}

\begin{figure}[tb]
    \centering
    \includegraphics[width=0.65\textwidth]{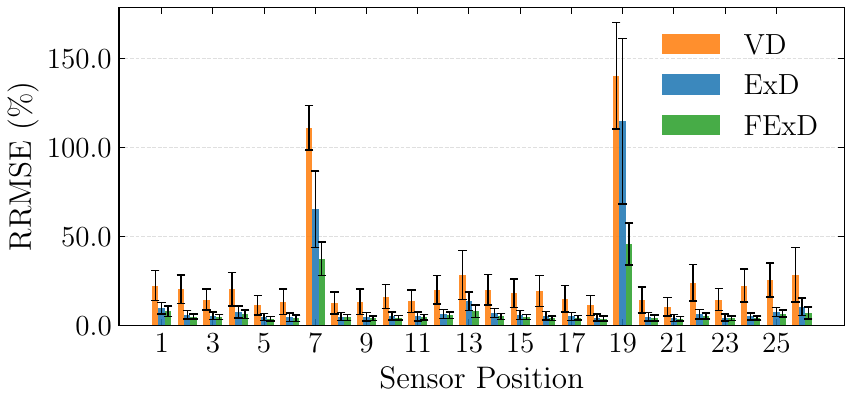}
    \caption{Bar plot for mean RRMSE errors across locations for different models with whiskers for one standard deviation.}\label{fig:error}
\end{figure}

While \Figz{fig:mae_hist}{fig:rrmse_hist} and \Tab{tab:error_metrics} provide performance results in the aggregate, it is also important to evaluate spatial performance.  This is particularly relevant given the bi-modal distributions for the RRMSE.  Therefore, 
\Fig{fig:error} separates the RRMSE by location, providing the mean value plus-or-minus one standard deviation of the RRMSE at each of the 26 sensor locations for each DeepONet model. Note that complete error metrics for all locations, including MAE and RMSE, can be found in \ref{sec:full_error}.

\begin{figure}[tb]
    \centering
    \begin{subfigure}{0.49\textwidth}
        \centering
        \includegraphics[width=\linewidth]{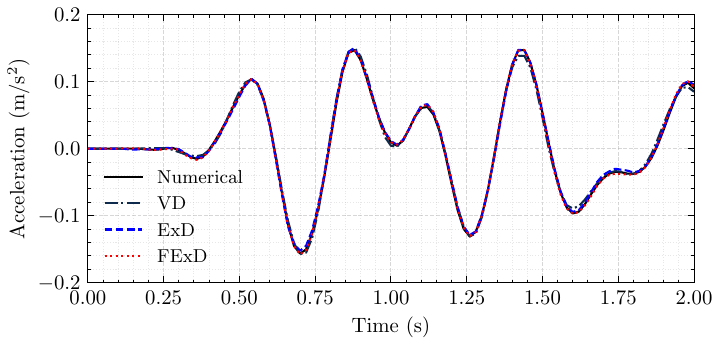}
        \caption{Position 5}\label{fig:pos_5}
    \end{subfigure}
    \hfill
    \begin{subfigure}{0.49\textwidth}
        \centering
        \includegraphics[width=\linewidth]{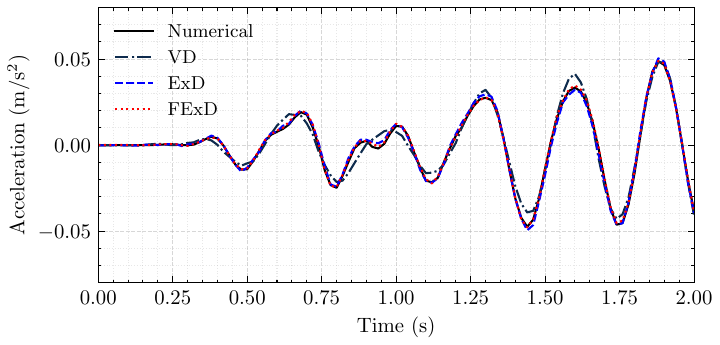}
        \caption{Position 13}\label{fig:pos_13}
    \end{subfigure}
    \caption{Prediction comparison of Full-field Ex-DeepONet with numerical responses at selected sensor positions.}
    \label{fig:mdof_2s_resp_pos_513}
\end{figure}

The majority of measurement locations exhibit reasonable prediction accuracy across all models, suggesting all of the considered DeepONets can approximately represent the observed structural behavior. For instance, \Fig{fig:mdof_2s_resp_pos_513} compares predictions from all three models against the numerical responses at two selected sensor locations, 5 and 13. Sensor location 5 is chosen because it exhibits the lowest RRMSE among all locations, and \Fig{fig:pos_5} illustrates that all models deliver high-quality response predictions.  Sensor location 13 corresponds to the mid-span of the bridge deck; thus, it is an important position from a dynamics perspective.  \Figu{fig:pos_13} shows that all three methods still provide quality estimes, but the prediction for the FExD is essentially co-linear and, thus, indistinguishable from the true response. At both locations, the FExD demonstrates superior agreement with target responses compared to VD and ExD. 

\begin{figure}[tb]
	\centering
	\begin{subfigure}{0.49\textwidth}
		\centering
		\includegraphics[width=\linewidth]{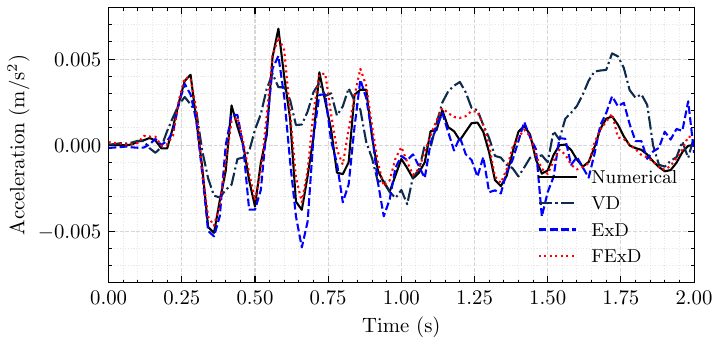}
        \caption{Position 7}\label{fig:pos_7}
	\end{subfigure}
    \hfill
     \begin{subfigure}{0.49\textwidth}
		\centering
		\includegraphics[width=\linewidth]{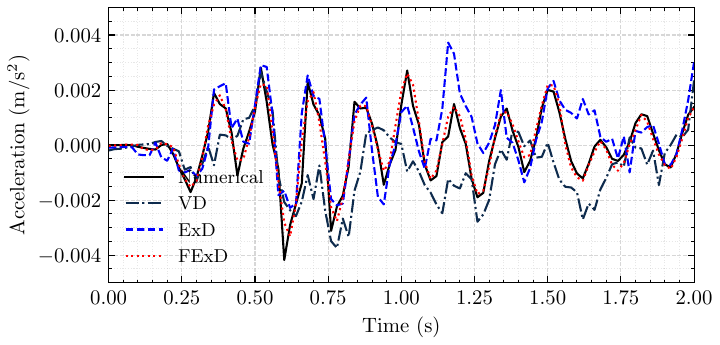}
        \caption{Position 19}\label{fig:pos_19}
	\end{subfigure}
	\caption{Prediction comparison of Full-field Ex-DeepONet with numerical responses with 2s.}
	\label{fig:mdof_fexd_2s_resp_pos_719}
\end{figure}

However, \Fig{fig:error} also shows that there are some sensor positions with rather large errors due to their proximity to structural discontinuities or regions with complex behavior, such as the side span of the bridge deck. 
For instance,  mean RRMSE values of approximately 37.48\% and 45.78\% are observed for even the FExD model at positions 7 and 19. These higher errors are due to the small response magnitudes near the piers that amplify relative error metrics. \Fig{fig:mdof_fexd_2s_resp_pos_719} presents a comparison of time-history predictions at these two positions. Notably, their response amplitudes are approximately one order-of-magnitude lower than those at positions 5 and 13, which means that while these locations provide minor absolute errors, they magnify relative error metrics.
\Figu{fig:mdof_fexd_2s_resp_pos_719} clearly demonstrates that both the VD and ExD struggle with these low amplitude predictions, but
the FExD model still captures the temporal patterns with reasonable fidelity by maintaining consistent magnitude and phase alignment throughout the time series. 

This is a particularly significant result because the challenge posed at these locations represents a fundamental issue in structural dynamics surrogate modeling: the need to simultaneously represent responses across locations with dramatically different magnitudes within a single architecture. While positions with larger responses benefit from stronger signal characteristics, positions near structural discontinuities must capture subtle, small-scale dynamics.  This is evidently a weakness of the VD and ExD, but the FExD still manages to provide dynamically consistent predictions, even if the accuracy is lower than at the high amplitude response locations.

\begin{figure}[tb]
	\centering
	\begin{subfigure}{\textwidth}
		\centering
		\includegraphics[width=\linewidth]{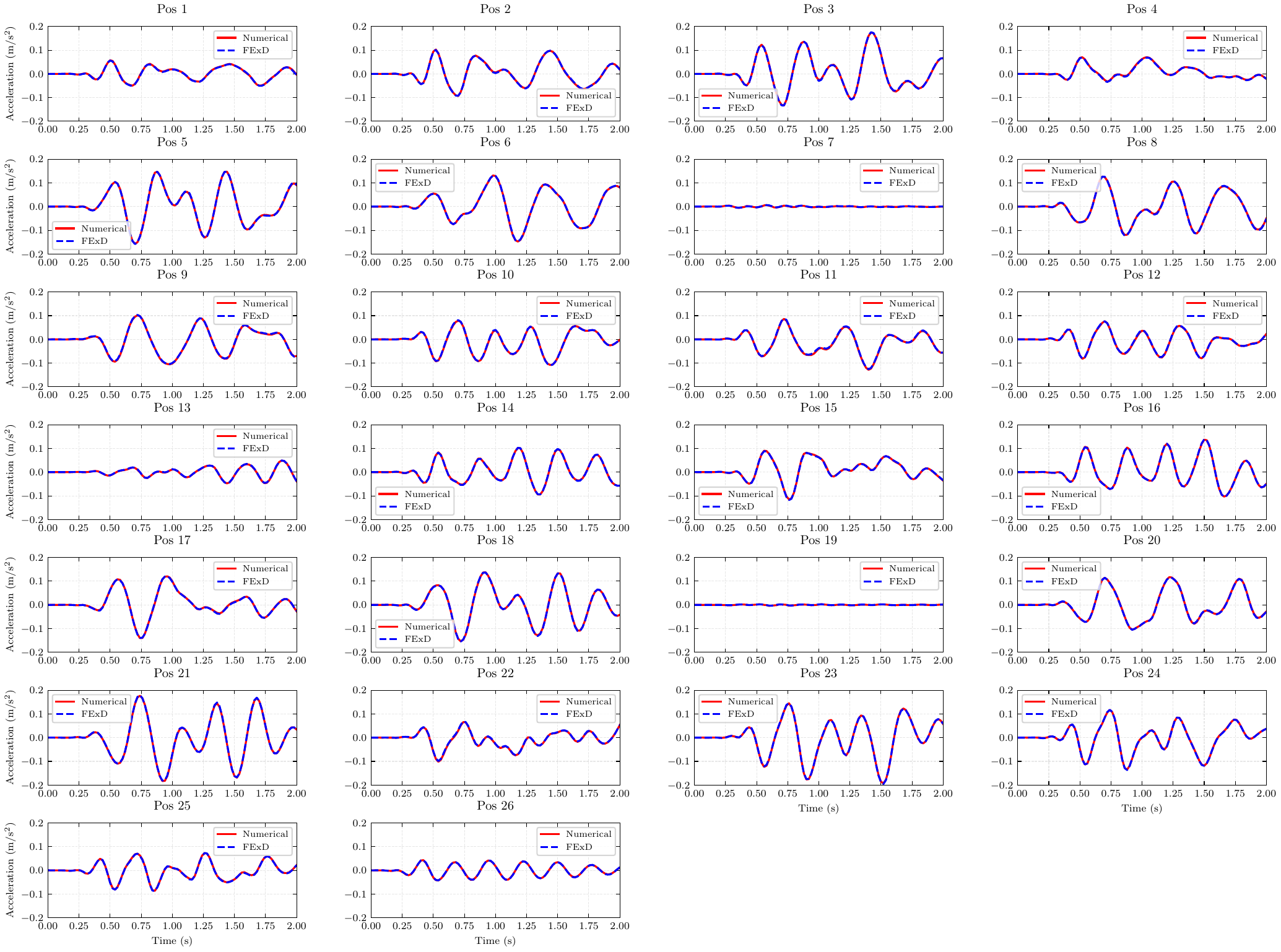}
	\end{subfigure}\\
	\caption{Comparison of numerical responses with predicted responses from the FExD across all 26 locations.}
	\label{fig:mdof_fexd_2s_resp}
\end{figure}

Lastly, the FExD demonstrates notably reduced error variability across sensor positions in \Fig{fig:error}, reflecting greater robustness and consistency under varying input conditions.  For instance, the standard deviation among the mean RRMSE values, \ie the standard deviation of the bar heights, is only 10.10\% for the FExD, compared to 23.81\% for the ExD and 30.04\% for the VD considering all 26 locations.  Removing the larger RRMSE values at locations 7 and 19 significantly lowers the standard deviation to merely 1.29\% for the FExD.  
This enhanced stability, coupled with improved accuracy, highlights the FExD's promise as a reliable surrogate model for predicting dynamical structural responses at numerous spatial locations.  This capacity of the FExD is illustrated in \Fig{fig:mdof_fexd_2s_resp}, which presents a sample of the full-field response prediction from one of the test cases.

\subsubsection{Worst case scenario}

\begin{figure}[tb]
	\centering
	\begin{subfigure}{0.49\textwidth}
		\centering
		\includegraphics[width=\linewidth]{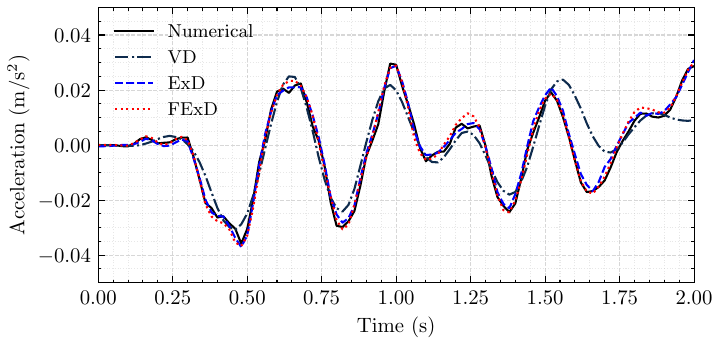}
        \caption{Position 5}\label{fig:worst_2s_pos5}
	\end{subfigure}
    \hfill
     \begin{subfigure}{0.49\textwidth}
		\centering
		\includegraphics[width=\linewidth]{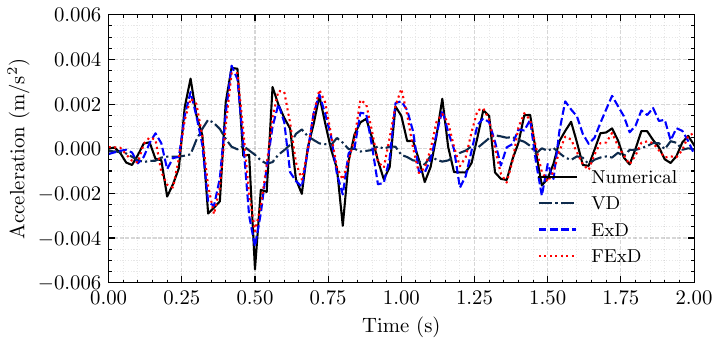}
        \caption{Position 7}\label{fig:worst_2s_pos7}
	\end{subfigure}
	\caption{Comparison of response predictions at sensor positions (a) 5 and (b) 7 for the ``worst case'' from testing for the FExD.}
	\label{fig:mdof_fexd_2s_resp_worst}
\end{figure}

A surrogate model is only as good, and as trustworthy, as its ``worst'' case prediction.  \Figu{fig:mdof_fexd_2s_resp_worst} presents a comparison of the response predictions at two measurement locations, 5 and 7, for the ``worst'' test case for the FExD as measured by average RRMSE across all 26 locations.  Note that this case, realization number 108 for the test cases, is the worst case for the FExD and VD model; however, the ExD experiences its worst performance during a different test case, realization number 495.  The RRMSE at sensor position 5 is 12.08\% for the FExD, while it rises to 47.29\% at position 7.  However, even at these elevated levels, both \Figs{fig:worst_2s_pos5}{fig:worst_2s_pos7} demonstrate that the FExD still delivers high-quality predictions, including at difficult locations like sensor position 7.  Further, the worst case predictions from the FExD are also noticeably more stable than those for the VD and ExD models, as the FExD does not exhibit the drift in its predictions observed for the VD and ExD models in \Fig{fig:worst_2s_pos7} and previously in \Fig{fig:mdof_fexd_2s_resp_pos_719}.

\Tab{tab:error_worst} reports the RRMSE values at sensor positions 5 and 7 as well as the average RRMSE error the FExD's worst case scenario (realization 108).  As previously mentioned, this is not the worst case for the ExD model; thus, it performs slightly better at sensor position 5 and has a comparable mean error. However, \Tab{tab:error_worst} also features the error metrics from realization 495, the worst case for the ExD model.  This shows that the worst case for the ExD leads its total RRMSE to rise to 41.54\%, which is well above the FExD's worst case.  Predictably, the VD model consistently provides large errors in both instances.  Thus, \Tab{tab:error_worst} demonstrates that even in worst case scenarios, the FExD consistently provides better, if not significantly, better results than the other DeepONets.

\begin{table}[tb]
    \centering
    \scriptsize
    \caption{Performance in terms of overall RRMSE for ``worst case'' scenarios}
    \label{tab:error_worst}
    \begin{tabular}{l|ccc|ccc}
    \toprule
    \multirow{2}{*}{Model}
    & \multicolumn{3}{c}{Worst VD/ FExD (Realization No. 108)} 
    & \multicolumn{3}{c}{Worst ExD (Realization No. 495)} \\
    & Position 5 & Position 7 & Overall (Mean) 
    & Position 5 & Position 7 & Overall (Mean)\\
    \midrule
    VD & $48.36 \%$ & $107.94 \%$ & $61.61 \%$ & $9.19 \%$ & $120.76 \%$ & $23.14\%$\\
    ExD & $11.82 \%$ & $60.26 \%$ & $17.83 \%$ & $9.03 \%$ & $184.55 \%$ & $41.54\%$\\
    FExD & $12.08 \%$ & $47.29 \%$ & $17.00 \%$ & $4.35 \%$ & $29.72 \%$ & $7.39\%$\\
    \bottomrule
    \end{tabular}
\end{table}

\subsubsection{Effect of training data size}

\begin{figure}
		\centering
		\includegraphics[width=0.4\linewidth]{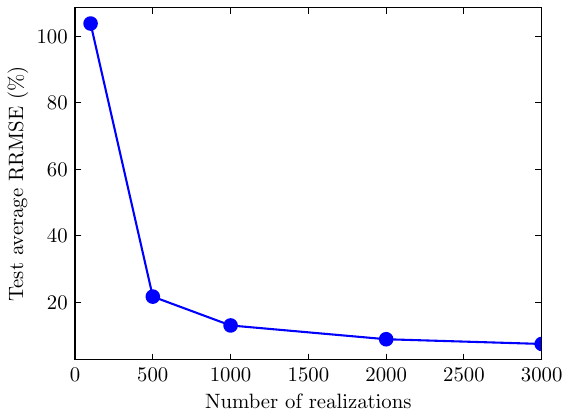}
		\caption{Performance vs. number of realizations for FExD.} \label{fig:rrmse_vs_nsim}
\end{figure}

\Figu{fig:rrmse_vs_nsim} plots the performance of the FExD, as measured by the average RRMSE in the test data, against the training data size.  The plot shows that the performance of the FExD starts to converge once 1000 realizations are included in the training data.  The final convergence can be seen when including 2000-3000 realizations in the training data.

\subsection{FExD on 10s data}

To explore the model performance of the proposed FExD on more complex temporal data, we retrained the model for 10-s simulations.  The bridge model is still subjected to stochastic ground motions generated from the same K-T spectrum and envelope function as stated in \Sec{sec:input}, but the duration is set to $t_n = 10$~s.  The same architecture as listed in  \Tab{tab:DeepONet_prm} is employed, but the batch size is increased to 1024 to accelerate training. The model converged after 2155 epochs in approximately 3.52 hours. Considering the impact that the extended temporal domain and increased data volume will have on training time, the VD and ExD are not trained for this investigation.  Also, note that training the FExD to learn the 10-s responses still requires only 20-25\% of the time it takes to train the VD and ExD on 2-s response data.

\Tab{tab:error_metrics_fexd_10s} presents the mean values plus-or-minus one standard deviation for the four error metrics across all 26 measurement positions. The FExD model maintains strong performance for most positions, with notable exceptions at locations 7 and 19, as expected.  Most measurement locations showcase an excellent data fit with $R^2$ values exceeding 95\% at every sensor location other than 7, 19, and 26.  In total, 10 of the 26 locations, or nearly 40\% of sensor locations, have $R^2$ values above 98\%.  Similarly, the RRMSE values are below 20\% at 21 of the 26 locations, with the piers (7 and 19), endpoints (1 and 26), and mid-span (13) providing the only exceptions.  The errors at locations 7 and 19 are particularly notable as the their mean RRMSE values are 97.01\% and 89.54\%, respectively, while their mean $R^2$ scores are 5.78 and 19.58.  However, these poor metrics are essentially outliers compared to the other locations.

The bottom row of \Tab{tab:error_metrics_fexd_10s} reports the average error metrics across locations, showing an average MAE of $0.0058$~m/s$^2$, average RRMSE of 21.81\%, and average $R^2$ value of 90.65\%. 
The MAE, RMSE, and RRMSE metrics all exceed those shown for the FExD in \Tab{tab:error_metrics}; however, comparing these two tables reveals that the FExD with 10-s predictions still outperforms the VD for 2-s predictions in terms of RRMSE.  In addition, the mean and median $R^2$ values in \Tab{tab:error_metrics_fexd_10s} are noticeably lower than those shown for the FExD in \Tab{tab:error_metrics}, but the standard deviations are still comparable, suggesting that the FExD maintains its performance consistency.  Further, the FExD with 10-s predictions once again surpasses the VD for 2-s predictions in terms of $R^2$ metrics.

\begin{table}[tb]
\centering
\scriptsize
\caption{Error metrics across sensor positions for FExD model considering 10~s dynamical response predictions.}
\label{tab:error_metrics_fexd_10s}
\begin{tabular}{l|ccc|ccc|ccc|ccc}
\toprule
\multirow{2}{*}{Position} & \multicolumn{3}{c|}{MAE (m/s$^2$)} & \multicolumn{3}{c|}{RMSE (m/s$^2$)} & \multicolumn{3}{c|}{RRMSE (\%)} & \multicolumn{3}{c}{R$^2$ (\%)} \\
\cline{2-13}
& Mean & Median & Std & Mean & Median & Std  & Mean & Median & Std & Mean & Median & Std  \\
\midrule
1  & 0.0041 & 0.0041 & 0.0007 & 0.0055 & 0.0054 & 0.0009 & 20.14 & 20.01 & 2.66 & 95.87 & 95.99 & 1.10 \\
2  & 0.0060 & 0.0059 & 0.0012 & 0.0079 & 0.0077 & 0.0015 & 16.02 & 15.77 & 2.43 & 97.37 & 97.51 & 0.81 \\
3  & 0.0074 & 0.0072 & 0.0016 & 0.0096 & 0.0094 & 0.0020 & 13.50 & 13.34 & 2.16 & 98.13 & 98.22 & 0.61 \\
4  & 0.0055 & 0.0053 & 0.0012 & 0.0070 & 0.0069 & 0.0015 & 14.13 & 13.74 & 3.25 & 97.89 & 98.11 & 1.01 \\
5  & 0.0070 & 0.0067 & 0.0016 & 0.0090 & 0.0088 & 0.0020 & 12.83 & 12.70 & 2.38 & 98.30 & 98.38 & 0.64 \\
6  & 0.0066 & 0.0064 & 0.0013 & 0.0086 & 0.0084 & 0.0017 & 15.22 & 14.76 & 3.52 & 97.56 & 97.82 & 1.16 \\
7  & 0.0012 & 0.0012 & 0.0002 & 0.0018 & 0.0018 & 0.0003 & 97.01 & 97.37 & 3.11 & 5.78 & 5.18 & 6.01 \\
8  & 0.0064 & 0.0063 & 0.0013 & 0.0084 & 0.0082 & 0.0017 & 14.47 & 14.28 & 2.84 & 97.82 & 97.96 & 0.87 \\
9  & 0.0063 & 0.0062 & 0.0015 & 0.0082 & 0.0080 & 0.0019 & 12.70 & 12.40 & 2.48 & 98.33 & 98.46 & 0.68 \\
10 & 0.0065 & 0.0064 & 0.0015 & 0.0085 & 0.0083 & 0.0019 & 12.82 & 12.63 & 2.67 & 98.28 & 98.40 & 0.73 \\
11 & 0.0067 & 0.0066 & 0.0017 & 0.0087 & 0.0084 & 0.0020 & 12.13 & 11.86 & 2.86 & 98.44 & 98.59 & 0.78 \\
12 & 0.0057 & 0.0056 & 0.0011 & 0.0075 & 0.0073 & 0.0014 & 16.30 & 16.25 & 2.67 & 97.27 & 97.36 & 0.90 \\
13 & 0.0042 & 0.0042 & 0.0007 & 0.0056 & 0.0055 & 0.0009 & 24.80 & 23.87 & 5.79 & 93.52 & 94.30 & 3.13 \\
14 & 0.0057 & 0.0056 & 0.0011 & 0.0075 & 0.0073 & 0.0014 & 16.44 & 16.20 & 2.53 & 97.23 & 97.38 & 0.86 \\
15 & 0.0066 & 0.0064 & 0.0016 & 0.0085 & 0.0082 & 0.0020 & 12.94 & 12.53 & 3.09 & 98.23 & 98.43 & 0.90 \\
16 & 0.0067 & 0.0065 & 0.0016 & 0.0087 & 0.0085 & 0.0020 & 12.83 & 12.61 & 2.62 & 98.28 & 98.41 & 0.70 \\
17 & 0.0064 & 0.0063 & 0.0015 & 0.0083 & 0.0082 & 0.0019 & 13.61 & 13.35 & 2.40 & 98.09 & 98.22 & 0.70 \\
18 & 0.0065 & 0.0063 & 0.0014 & 0.0085 & 0.0083 & 0.0018 & 14.23 & 13.89 & 2.32 & 97.92 & 98.07 & 0.70 \\
19 & 0.0006 & 0.0006 & 0.0001 & 0.0009 & 0.0009 & 0.0001 & 89.54 & 89.36 & 4.82 & 19.58 & 20.11 & 8.65 \\
20 & 0.0066 & 0.0065 & 0.0013 & 0.0087 & 0.0086 & 0.0017 & 15.64 & 15.28 & 3.10 & 97.46 & 97.67 & 1.04 \\
21 & 0.0075 & 0.0073 & 0.0017 & 0.0097 & 0.0095 & 0.0022 & 12.71 & 12.51 & 2.17 & 98.34 & 98.43 & 0.58 \\
22 & 0.0063 & 0.0062 & 0.0014 & 0.0082 & 0.0080 & 0.0017 & 15.01 & 14.57 & 3.02 & 97.65 & 97.87 & 0.97 \\
23 & 0.0074 & 0.0073 & 0.0017 & 0.0097 & 0.0096 & 0.0021 & 13.03 & 12.87 & 2.09 & 98.26 & 98.34 & 0.57 \\
24 & 0.0065 & 0.0064 & 0.0014 & 0.0085 & 0.0084 & 0.0017 & 15.25 & 15.13 & 2.25 & 97.62 & 97.71 & 0.71 \\
25 & 0.0052 & 0.0051 & 0.0009 & 0.0068 & 0.0067 & 0.0012 & 19.23 & 19.11 & 2.66 & 96.23 & 96.35 & 1.06 \\
26 & 0.0043 & 0.0043 & 0.0007 & 0.0059 & 0.0058 & 0.0009 & 34.50 & 34.02 & 7.20 & 87.58 & 88.43 & 5.28 \\
\midrule
\textbf{Average} & \textbf{0.0058} & \textbf{0.0056} & \textbf{0.0012} & \textbf{0.0076} & \textbf{0.0074} & \textbf{0.0016} & \textbf{21.81} & \textbf{21.55} & \textbf{3.04} & \textbf{90.65} & \textbf{90.83} & \textbf{1.58} \\
\bottomrule
\end{tabular}
\end{table}

\begin{figure}[htb]
    \centering
    \begin{subfigure}[b]{0.3\textwidth}
        \centering
        \includegraphics[width=\linewidth]{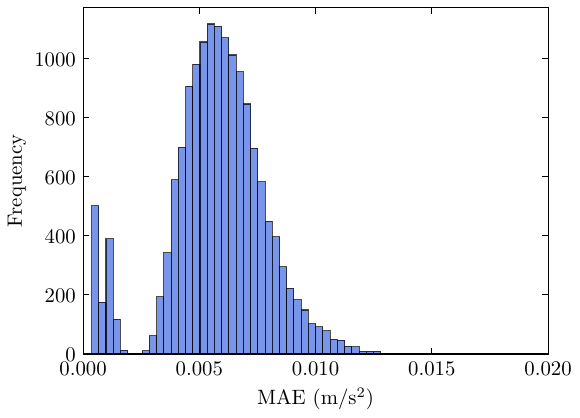}
        \caption{Testing MAE for FExD}
        \label{fig:fexd_10s_mae_hist}
    \end{subfigure}
    \hspace{0.02\textwidth}
    \begin{subfigure}[b]{0.3\textwidth}
        \centering
        \includegraphics[width=\linewidth]{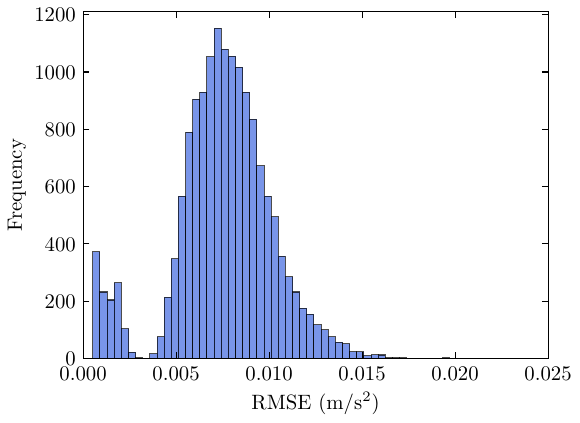}
        \caption{Testing RMSE for FExD}
        \label{fig:fexd_10s_rmse_hist}
    \end{subfigure}
        \hspace{0.02\textwidth}
    \begin{subfigure}[b]{0.3\textwidth}
        \centering
        \includegraphics[width=\linewidth]{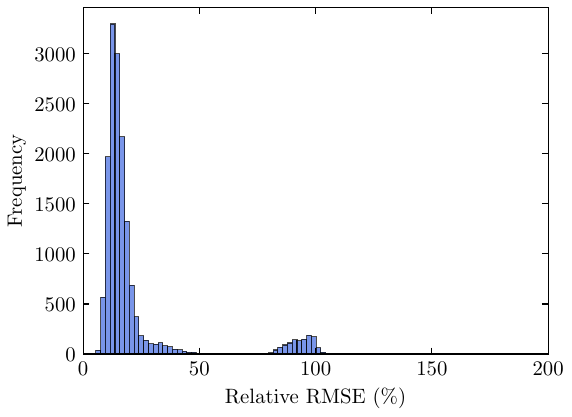}
        \caption{Testing RRMSE for FExD}
        \label{fig:fexd_10s_rrmse_hist}
    \end{subfigure}
    \caption{Comparison of testing error for the chosen spatial-temporal DeepONets with 10s.}
    \label{fig:mdof_fexd_hist_10s}
\end{figure}

\Fig{fig:mdof_fexd_hist_10s} provides histograms of the error metrics considering all measurement locations. The histograms demonstrate the model consistency, with RRMSE values (\Fig{fig:fexd_10s_rrmse_hist}) predominantly clustered below 40\%, indicating great relative accuracy for most predictions. The absolute error metrics MAE (\Fig{fig:fexd_10s_mae_hist}) and RMSE (\Fig{fig:fexd_10s_rmse_hist}) show distributions primarily concentrated below $0.01$~m/s$^2$, reflecting the model's ability to maintain precision even over extended time periods.  All three histograms also exhibit bi-modality.  While the bi-modality of the RRMSE is expected based on previous results, the small grouping of error values near zero for both the MAE and RMSE was not observed for the 2-s response predictions.  

The bi-modality of the MAE and RMSE plots occurs because the majority of errors are larger than for the 2-s case, pushing the distributions away from zero.  However, there are still a few locations that, due to their small response amplitudes given close proximity to piers, produce small errors that tend towards zero, \ie sensor positions 7 and 19.  In fact, sensor positions 7 and 9 account for over 94\% of the ``mass'' within the small groupings near zero error shown in \Figs{fig:fexd_10s_mae_hist}{fig:fexd_10s_rmse_hist}.  The remainder comes from  locations that happen to produce low error estimates during a few realizations.

\begin{figure}[tb]
    \centering
    \begin{subfigure}{0.45\textwidth}
        \centering
        \includegraphics[width=\linewidth]{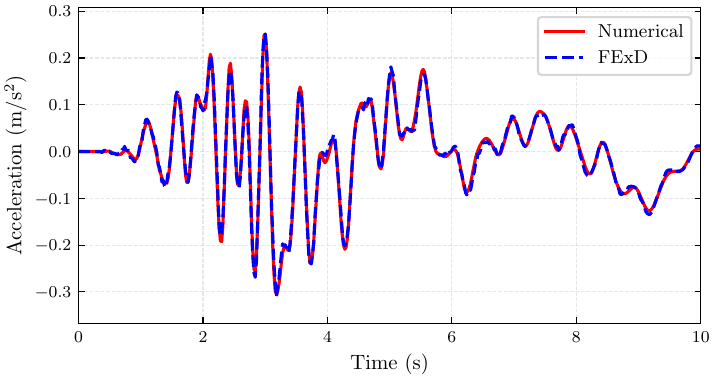}
        \caption{Position 5}\label{fig:pos_5_10s}
    \end{subfigure}
    \hfill
    \begin{subfigure}{0.45\textwidth}
        \centering
        \includegraphics[width=\linewidth]{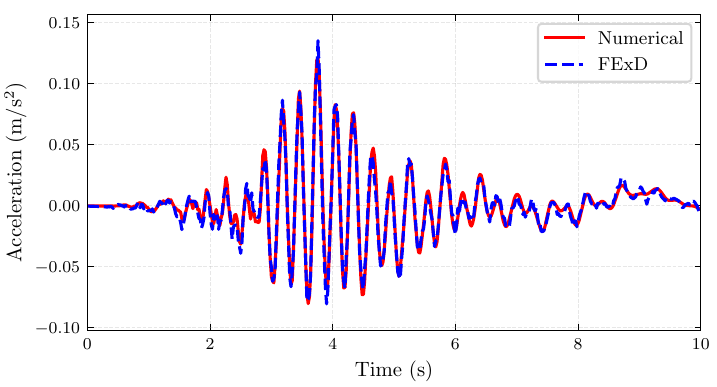}
        \caption{Position 13}\label{fig:pos_13_10s}
    \end{subfigure}\\
    \vspace{9pt}
	\begin{subfigure}{0.45\textwidth}
		\centering
		\includegraphics[width=\linewidth]{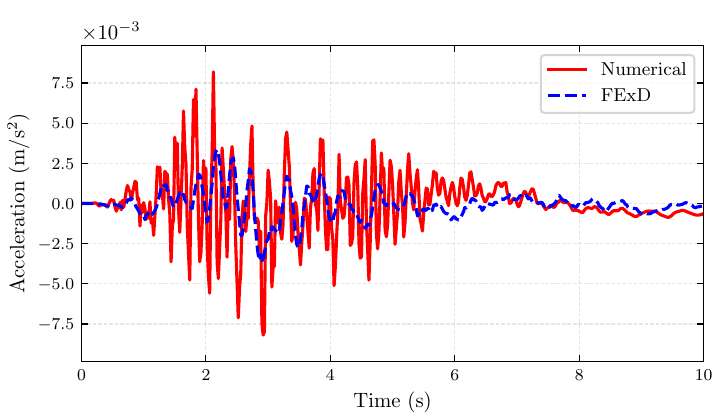}
        \caption{Position 7}\label{fig:pos_7_10s}
	\end{subfigure}
    \hfill
     \begin{subfigure}{0.45\textwidth}
		\centering
		\includegraphics[width=\linewidth]{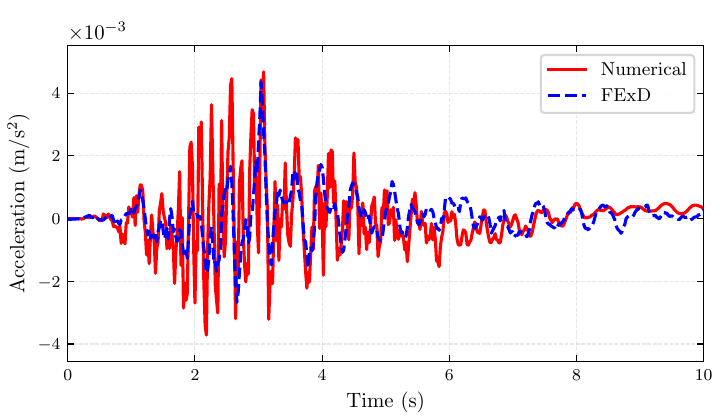}
        \caption{Position 19}\label{fig:pos_19_10s}
	\end{subfigure}
	\caption{Comparison of FExD predictions at sensor positions (a) 5, (b) 12, (c) 7, and (d) 19 when trained on 10~s of response data.}
	\label{fig:mdof_fexd_10s_pos_513719}
\end{figure}

\Figu{fig:mdof_fexd_10s_pos_513719} presents response predictions from the FExD at a few specified sensor locations for a sample test case.  The predicted time histories at locations 5 and 13 are quite accurate, as they were when the FExD was trained on 2~s of response data.  This provides further evidence of the ability of the FExD to deliver high-quality predictions over long simulation, and thus excitation, time spans.  The plots for positions 7 and 19 are less encouraging, as the predictions disply the general trend in the response but failure to capture higher frequency oscillations.  Close inspection of the ordinate axis reveals that these responses are orders of magnitude smaller than those at positions 5 and 13; nonetheless, these predictions clearly fail to include all of the important dynamic characteristics.

\subsubsection{Frequency response prediction}

\begin{figure}[tb]
	\centering
		\includegraphics[width=0.45\linewidth]{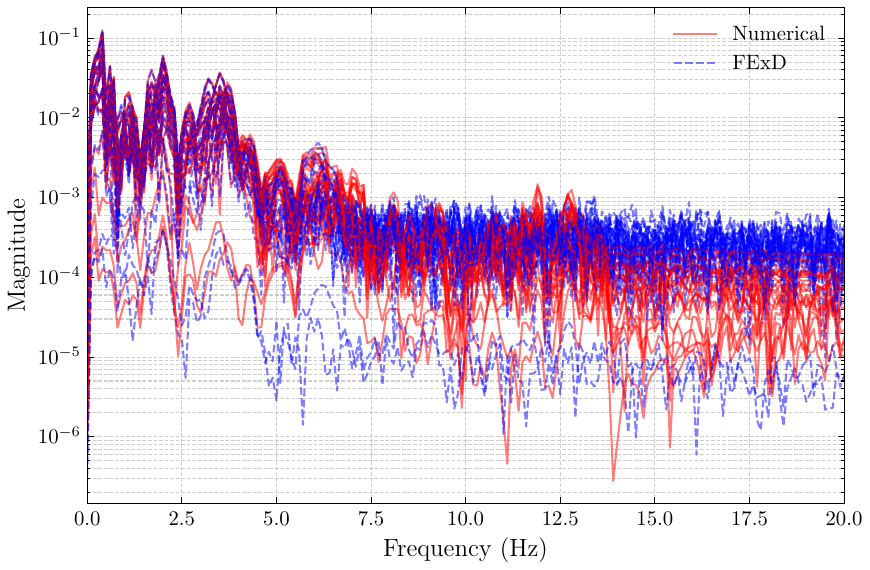}
		\caption{Fast Fourier Transform analysis comparing numerical and predicted responses.}
		\label{fig:fft_comparison}
\end{figure}

Properly capturing frequency content is an important aspect for demonstrating that a given surrogate is truly learning the dynamical structural system.  
\Figu{fig:fft_comparison} presents a Fast Fourier Transform (FFT) comparison of the numerical responses and FExD predictions across all 26 locations.  This figure demonstrates that the FExD captures the frequency information for the first several modes, as the spectra from the numerical responses and FExD show good agreement up to approximately 7.5~Hz.

\begin{figure}[tb]
	\centering
	\begin{subfigure}{\textwidth}
		\centering
		\includegraphics[width=0.45\linewidth]{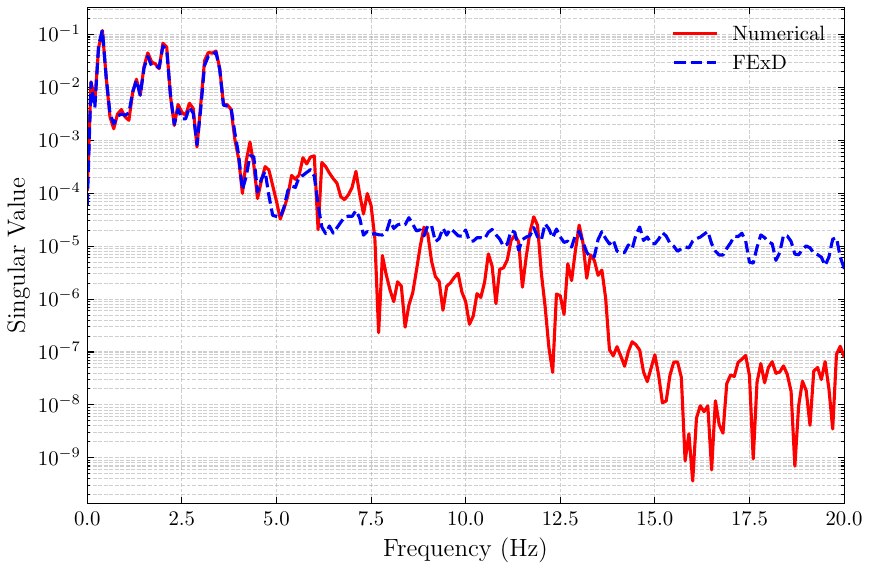}
		\label{fig:fdd_comparison}
	\end{subfigure}
	\caption{Singular value decomposition of the cross-power spectral density matrices derived from both numerical solutions and FExD predictions. 
    }
	\label{fig:frequency_comparison}
\end{figure}

The global frequency content of a system can be better considered through a singular value decomposition of the cross-power spectral density matrix.  This approach, as espoused by the frequency domain decomposition (FDD) method \cite{brincker_2001}, provides a better platform for viewing the ``global'' frequency content of the system. 
Despite difficulties predicting small-magnitude responses at some locations, \Fig{fig:frequency_comparison} emphasizes the strong alignment between the predicted and reference spectral characteristics, showing that the FExD successfully preserves the global frequency characteristics of the structural system. The excellent agreement between predicted and reference spectral signatures spans across the higher ``power'' portion of the frequency range, with particularly accurate identification of the first several dominant structural modes. The agreement in frequency content degrades after 5-6~Hz, but the singular value plot demonstrates that there is much less ``power'' in these frequencies even for the numerical model.  Thus, while local high-frequency oscillations might present challenges at specific low-magnitude positions, the FExD model can successfully preserve the essential modal properties and frequency characteristics critical for surrogate modeling of dynamical systems.

\subsubsection{Worst case scenario}

\begin{figure}[tb]
	\centering
	\begin{subfigure}{0.49\textwidth}
		\centering
		\includegraphics[width=\linewidth]{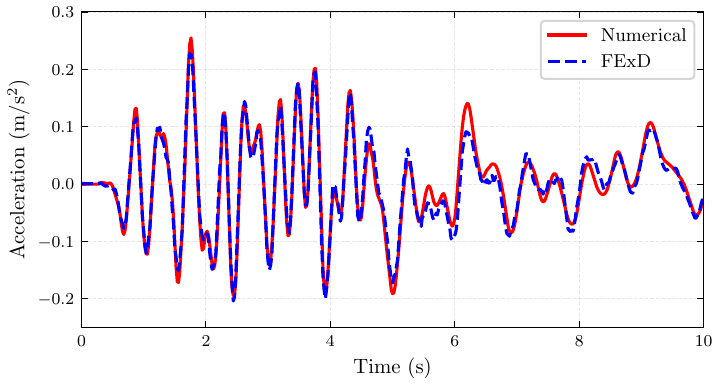}
        \caption{Position 5}\label{fig:worst_10s_pos5}
	\end{subfigure}
    \hfill
     \begin{subfigure}{0.49\textwidth}
		\centering
		\includegraphics[width=\linewidth]{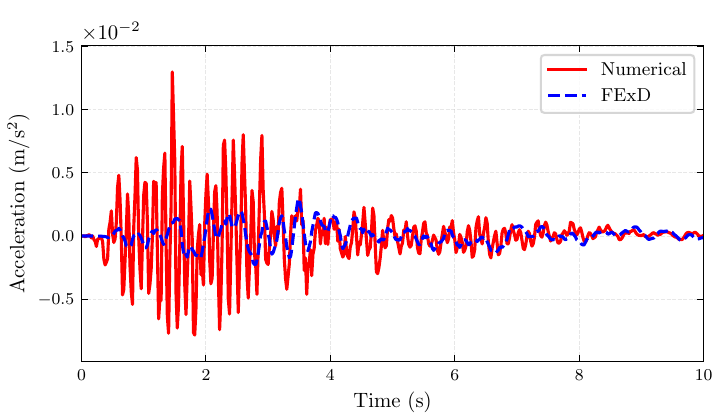}
        \caption{Position 7}\label{fig:worst_10s_pos7}
	\end{subfigure}
	\caption{Prediction comparison of Full-field Ex-DeepONet against numerical responses with 10s at worst cases.}
	\label{fig:mdof_fexd_10s_resp_worst}
\end{figure}

As with the 2-s simulations, it is critical to evaluate the performance of the FExD for the worst case among the 10-s simulations.  \Figu{fig:mdof_fexd_10s_resp_worst} presents the predictions at sensor locations 5 and 7 for the worst test case as measured by average RRMSE.  \Figu{fig:worst_10s_pos5} shows that the prediction at sensor location 5 generally follows the correct response trajectory, but several peak accelerations are underestimated.  These underestimations lead to an RRMSE at location 5 of 19.91\%, which is still reasonable albeit much larger than the RRMSE at the same location for the worst case among the 2-s simulations.  The predicted response at location 7 is quite poor, as the amplitude is much lower than the true response.  While the RRMSE for \Fig{fig:worst_10s_pos7} is over 98\%, this is actually in line with the general performance at this location, as \Tab{tab:error_metrics_fexd_10s} shows that the average RRMSE at this location is just above 97\%.  In contrast, the average error at location 5 is only 12.83\%, meaning that this worst case produces a 55\% relative increase in error.  This suggests that the ``difficult'' locations, \ie 7 and 19, do not contribute that much to the determination of the worst case as the FExD consistently struggled to provide reasonable predictions at these locations.

Overall, mean RRMSE for this worst case is 28.95\% as compared to the overall mean RRMSE of 21.81\% reported in \Tab{tab:error_metrics_fexd_10s}.  Thus, even in the worst case, %the FExD is still capable of providing reasonable estimates across all sensor locations.  
these metrics demonstrate that the FExD can effectively provide reasonable predictions of spatio-temporal responses for longer duration dynamical data, making it especially valuable for applications that require rapid assessments of structural performance under extended excitation scenarios.

\section{Conclusions}

This study introduces the full-field extended DeepONet (FExD) as a means for using operator-based learning to predict the full field of spatio-temporal responses from dynamical systems.  The FExD leverages the architecture of extended DeepONet (ExD) to maximize the expressiveness through enhanced basis formation and branch interactions while also utilizing the trunk to provide multiple dynamical outputs in a single pass.  The performance of the FExD is compared against a vanilla DeepONet (VD) as well as the conventional ExD.  However, to facilitate comparisons for multi-output cases, both the VD and ExD are modified to utilize spatio-temporal inputs to their trunk networks.  All three DeepONets, VD, ExD, and FExD, are then applied to learn and subsequently predict responses at 26 locations along the deck of a cable-stayed bridge model that served as a structural dynamics testbed.  

It was found that the FExD significantly outperformed both the VD and ExD during training, as it provided lower validation errors in much less time.  Due to the more expressive and efficient architecture of the FExD, it was able to complete more epochs in much less time, completing training nearly 15x's faster than the ExD and 23x's faster than the VD.  Further, the FExD outperformed both other DeepONets over every error metric in terms of mean and median value while also providing greater consistency through its narrower distributions, and subsequent smaller standard deviations.  The stability and consistency of the FExD was also illustrated when considering its location-by-location errors, as its predictions had the lowest variation, and its worst case scenario, which was markedly better than either of those for the ExD or VD.  

When considering responses at specific locations, the FExD demonstrated an ability to provide reasonably accurate predictions across orders of magnitudes, as it delivered quality predictions at both high amplitude locations between supports as well as at sensing locations close to the piers, where the response amplitudes are much lower.  When applied to longer duration simulations, the FExD still performed well, but its error metrics increased.  The increases were most significant at locations near the piers such that those predictions became quite poor.  However, the overall error performance for the longer duration simulations was still quite strong and the FExD was shown to capture most of the significant frequency content in the true responses.  

The other notable aspect about the longer duration responses was that the training times for both the VD and ExD become prohibitive, while the FExD completed its training in less time than the VD required for the short-duration simulations.  Thus, while the performance of the FExD has clear room for improvement, the proposed FExD architecture offers a new pathway for full-field dynamical response prediction that becomes impractical with traditional DeepONet architectures.  By inherently encoding spatial correlations, the FExD addresses a fundamental challenge in surrogate modeling of structural dynamics, where inter-dependencies between degrees of freedom significantly impact system behavior. 

The performance of the FExD represents a promising advancement in the development of neural operators as surrogates for engineering applications, particularly for efficient response prediction in structural dynamics scenarios.  Such rapid predictions could significantly influence and improve both optimization and design tasks where numerous simulations are typically required.  Future studies should consider methods to further improve the performance of the FExD for longer duration simulations beyond the inclusion of more training data.  For instance, DeepONets provide a versatile framework without architectural constraints on the branch and trunk networks; thus, futures studies could potentially move beyond standard MLPs and consider using emergent architectures such as transformers for the branch and/or trunk networks. Future improvements might include adaptive weighting schemes to compensate for magnitude disparities or hierarchical approaches that allocate additional modeling capacity to those challenging regions with higher frequency content.

\section*{CRediT Authorship Contribution Statement}

\textbf{Jichuan Tang}: Conceptualization, Methodology, Software, Investigation, Writing - Original Draft, Visualization. 
\textbf{Patrick T. Brewick}: Conceptualization, Methodology, Software, Investigation, Writing - Review \& Editing, Visualization, Supervision.\textbf{Ryan G. McClarren}: Methodology, Software, Supervision. 
\textbf{Christopher Sweet}: Methodology, Software, Supervision.

\section*{Declaration of Competing Interest}

The authors declare that they have no known competing financial interests or personal relationships that could have appeared to influence the work reported in this paper.

\section*{Acknowledgments}

The authors gratefully acknowledge the support of computational resources from the Center of Research Computing at Notre Dame towards this research.

\section*{Data availability}

Data will be made available on request.

\bibliographystyle{unsrt}
\bibliography{Refs.bib}

% \newpage
\appendix
\setcounter{table}{0}
\renewcommand{\thetable}{\Alph{section}\arabic{table}}
\setcounter{figure}{0}
\renewcommand{\thefigure}{\Alph{section}\arabic{figure}}
\setcounter{equation}{0}
\renewcommand{\theequation}{\Alph{section}\arabic{equation}}

\section{Modal Frequencies}

\Tab{tab:mod_freq} displays the natural frequencies identified from an eigenvalue analysis of the bridge model. Note that the frequencies listed in \Tab{tab:mod_freq} span the range from 0--20~Hz.
\begin{table}[ht]
    \centering
    \caption{Identified modal natural frequencies of the benchmark cable-stayed bridge}\label{tab:mod_freq}
    \scriptsize
    \begin{tabular}{c|c|c|c|c|c}
        \toprule
        Mode No. & Frequency (Hz) & Mode No. & Frequency (Hz) & Mode No. & Frequency (Hz) \\
        \midrule
         1 & 0.2899 &  2 & 0.3699 &  3 & 0.5812\\
         4 & 0.6490 &  5 & 0.7102 &  6 & 0.8508\\
         7 & 1.0013 &  8 & 1.0248 &  9 & 1.0684\\
        10 & 1.1109 & 11 & 1.2587 & 12 & 1.4521\\
        13 & 1.5926 & 14 & 1.6445 & 15 & 1.8877\\
        16 & 2.2084 & 17 & 2.4890 & 18 & 2.5778\\
        19 & 2.8091 & 20 & 3.4253 & 21 & 3.8314\\
        22 & 3.9319 & 23 & 4.2679 & 24 & 4.7732\\
        25 & 5.5687 & 26 & 5.7991 & 27 & 6.1844\\
        28 & 6.7118 & 29 & 7.9323 & 30 & 8.3648\\
        31 & 9.0105 & 32 & 9.5198 & 33 & 11.1870\\
        34 & 11.7730 & 35 & 12.6992 & 36 & 12.7097\\
        37 & 13.3893 & 38 & 15.3071 & 39 & 15.6810\\
        40 & 15.8349 & 41 & 18.6789 & 42 & 18.7668\\
        \bottomrule
    \end{tabular}
\end{table}

\section{Additional error metric reporting} \label{sec:full_error}
The full suite of evaluation metrics are presented location-by-location for each of the considered DeepONet models.  \Tab{tab:full_err_VD} provides the metrics for the VD model, \Tab{tab:full_err_ExD} provides the metrics for the ExD model, and \Tab{tab:full_err_FExD} provides the metrics for the FExD model.
% \subsection{2s data}
\begin{table}[htbp]
    \centering
    \scriptsize
    \caption{Error metrics across measurement positions for VD model} \label{tab:full_err_VD}
    \begin{tabular}{l|r|r|r|r}
    \toprule
    Position & MAE (m/s$^2$) & RMSE (m/s$^2$) & RRMSE (\%) & R$^2$ (\%) \\
    \midrule
    1 & $0.0035 \pm 0.0009$ & $0.0045 \pm 0.0011$ & $22.40 \pm 8.33$ & $94.25 \pm 4.89$ \\
    2 & $0.0056 \pm 0.0014$ & $0.0073 \pm 0.0018$ & $20.45 \pm 8.00$ & $95.14 \pm 4.41$ \\
    3 & $0.0057 \pm 0.0014$ & $0.0075 \pm 0.0018$ & $14.53 \pm 5.82$ & $97.53 \pm 2.41$ \\
    4 & $0.0042 \pm 0.0011$ & $0.0055 \pm 0.0014$ & $20.35 \pm 9.39$ & $94.76 \pm 5.72$ \\
    5 & $0.0043 \pm 0.0013$ & $0.0058 \pm 0.0017$ & $11.39 \pm 5.51$ & $98.38 \pm 2.05$ \\
    6 & $0.0041 \pm 0.0012$ & $0.0054 \pm 0.0015$ & $13.24 \pm 7.23$ & $97.71 \pm 3.12$ \\
    7 & $0.0016 \pm 0.0004$ & $0.0020 \pm 0.0005$ & $111.10 \pm 12.47$ & $-25.02 \pm 29.25$ \\
    8 & $0.0040 \pm 0.0010$ & $0.0052 \pm 0.0013$ & $12.65 \pm 6.13$ & $98.02 \pm 2.53$ \\
    9 & $0.0038 \pm 0.0013$ & $0.0052 \pm 0.0018$ & $13.39 \pm 7.21$ & $97.64 \pm 3.37$ \\
    10 & $0.0046 \pm 0.0012$ & $0.0059 \pm 0.0015$ & $16.20 \pm 6.66$ & $96.76 \pm 3.22$ \\
    11 & $0.0043 \pm 0.0012$ & $0.0055 \pm 0.0014$ & $13.51 \pm 6.26$ & $97.66 \pm 2.76$ \\
    12 & $0.0048 \pm 0.0012$ & $0.0063 \pm 0.0015$ & $20.05 \pm 7.92$ & $95.22 \pm 4.31$ \\
    13 & $0.0024 \pm 0.0007$ & $0.0032 \pm 0.0009$ & $28.26 \pm 13.77$ & $90.05 \pm 10.36$ \\
    14 & $0.0051 \pm 0.0014$ & $0.0067 \pm 0.0019$ & $20.16 \pm 8.61$ & $95.11 \pm 4.61$ \\
    15 & $0.0056 \pm 0.0017$ & $0.0072 \pm 0.0021$ & $18.08 \pm 7.92$ & $95.93 \pm 4.13$ \\
    16 & $0.0060 \pm 0.0019$ & $0.0077 \pm 0.0024$ & $19.34 \pm 8.85$ & $95.26 \pm 4.77$ \\
    17 & $0.0044 \pm 0.0014$ & $0.0061 \pm 0.0021$ & $15.07 \pm 7.52$ & $97.12 \pm 3.77$ \\
    18 & $0.0040 \pm 0.0012$ & $0.0053 \pm 0.0017$ & $11.20 \pm 5.63$ & $98.43 \pm 2.10$ \\
    19 & $0.0010 \pm 0.0002$ & $0.0013 \pm 0.0003$ & $140.40 \pm 29.94$ & $-106.19 \pm 94.44$ \\
    20 & $0.0043 \pm 0.0013$ & $0.0057 \pm 0.0016$ & $14.30 \pm 7.31$ & $97.41 \pm 3.24$ \\
    21 & $0.0044 \pm 0.0014$ & $0.0060 \pm 0.0020$ & $10.43 \pm 5.28$ & $98.62 \pm 1.81$ \\
    22 & $0.0062 \pm 0.0020$ & $0.0079 \pm 0.0024$ & $23.98 \pm 10.33$ & $93.00 \pm 6.72$ \\
    23 & $0.0062 \pm 0.0017$ & $0.0080 \pm 0.0022$ & $14.49 \pm 6.22$ & $97.50 \pm 2.53$ \\
    24 & $0.0073 \pm 0.0024$ & $0.0093 \pm 0.0030$ & $22.36 \pm 9.27$ & $94.11 \pm 5.27$ \\
    25 & $0.0053 \pm 0.0016$ & $0.0067 \pm 0.0020$ & $25.46 \pm 9.58$ & $92.57 \pm 6.02$ \\
    26 & $0.0030 \pm 0.0009$ & $0.0038 \pm 0.0011$ & $28.47 \pm 15.44$ & $89.51 \pm 13.19$ \\
    \midrule
    \textbf{Average} & $\mathbf{0.0045 \pm 0.0013}$ & $\mathbf{0.0058 \pm 0.0017}$ & $\mathbf{26.20 \pm 9.10}$ & $\mathbf{83.33 \pm 8.88}$ \\
    \bottomrule
    \end{tabular}
\end{table}

\begin{table}[htbp]
\centering
\scriptsize
\caption{Error metrics across measurement positions for ExD model} \label{tab:full_err_ExD}
    \begin{tabular}{l|r|r|r|r}
    \toprule
    Position & MAE (m/s$^2$) & RMSE (m/s$^2$) & RRMSE (\%) & R$^2$ (\%) \\
    \midrule
    1 & $0.0015 \pm 0.0006$ & $0.0020 \pm 0.0008$ & $9.59 \pm 3.29$ & $98.96 \pm 0.88$ \\
    2 & $0.0017 \pm 0.0008$ & $0.0022 \pm 0.0010$ & $5.99 \pm 2.30$ & $99.59 \pm 0.46$ \\
    3 & $0.0022 \pm 0.0009$ & $0.0029 \pm 0.0012$ & $5.35 \pm 2.08$ & $99.67 \pm 0.38$ \\
    4 & $0.0017 \pm 0.0008$ & $0.0022 \pm 0.0010$ & $7.54 \pm 3.48$ & $99.28 \pm 0.97$ \\
    5 & $0.0019 \pm 0.0009$ & $0.0025 \pm 0.0011$ & $4.69 \pm 1.96$ & $99.74 \pm 0.32$ \\
    6 & $0.0016 \pm 0.0008$ & $0.0020 \pm 0.0010$ & $4.69 \pm 2.37$ & $99.72 \pm 0.40$ \\
    7 & $0.0009 \pm 0.0004$ & $0.0012 \pm 0.0004$ & $65.29 \pm 21.45$ & $52.76 \pm 39.41$ \\
    8 & $0.0017 \pm 0.0008$ & $0.0022 \pm 0.0010$ & $4.96 \pm 2.20$ & $99.70 \pm 0.36$ \\
    9 & $0.0015 \pm 0.0008$ & $0.0020 \pm 0.0011$ & $4.80 \pm 2.35$ & $99.71 \pm 0.42$ \\
    10 & $0.0016 \pm 0.0008$ & $0.0020 \pm 0.0010$ & $5.31 \pm 2.34$ & $99.64 \pm 0.62$ \\
    11 & $0.0017 \pm 0.0010$ & $0.0022 \pm 0.0013$ & $5.07 \pm 2.53$ & $99.66 \pm 0.65$ \\
    12 & $0.0016 \pm 0.0007$ & $0.0021 \pm 0.0009$ & $6.49 \pm 2.50$ & $99.50 \pm 0.68$ \\
    13 & $0.0012 \pm 0.0004$ & $0.0016 \pm 0.0004$ & $13.54 \pm 5.15$ & $97.88 \pm 1.76$ \\
    14 & $0.0019 \pm 0.0007$ & $0.0024 \pm 0.0008$ & $6.92 \pm 2.47$ & $99.45 \pm 0.58$ \\
    15 & $0.0018 \pm 0.0010$ & $0.0024 \pm 0.0013$ & $5.71 \pm 2.54$ & $99.59 \pm 0.69$ \\
    16 & $0.0018 \pm 0.0008$ & $0.0023 \pm 0.0011$ & $5.44 \pm 2.42$ & $99.63 \pm 0.62$ \\
    17 & $0.0017 \pm 0.0009$ & $0.0022 \pm 0.0012$ & $5.00 \pm 2.38$ & $99.69 \pm 0.47$ \\
    18 & $0.0017 \pm 0.0009$ & $0.0022 \pm 0.0011$ & $4.40 \pm 2.06$ & $99.76 \pm 0.34$ \\
    19 & $0.0008 \pm 0.0004$ & $0.0010 \pm 0.0005$ & $114.81 \pm 46.54$ & $-53.56 \pm 230.77$ \\
    20 & $0.0015 \pm 0.0008$ & $0.0020 \pm 0.0010$ & $4.72 \pm 2.38$ & $99.72 \pm 0.43$ \\
    21 & $0.0019 \pm 0.0009$ & $0.0025 \pm 0.0012$ & $4.16 \pm 1.92$ & $99.79 \pm 0.30$ \\
    22 & $0.0017 \pm 0.0009$ & $0.0022 \pm 0.0011$ & $6.25 \pm 2.80$ & $99.52 \pm 0.70$ \\
    23 & $0.0020 \pm 0.0010$ & $0.0025 \pm 0.0013$ & $4.36 \pm 2.00$ & $99.77 \pm 0.35$ \\
    24 & $0.0017 \pm 0.0008$ & $0.0022 \pm 0.0011$ & $5.01 \pm 2.08$ & $99.70 \pm 0.38$ \\
    25 & $0.0016 \pm 0.0007$ & $0.0020 \pm 0.0008$ & $7.44 \pm 2.52$ & $99.38 \pm 0.55$ \\
    26 & $0.0012 \pm 0.0006$ & $0.0015 \pm 0.0007$ & $10.43 \pm 4.93$ & $98.67 \pm 1.84$ \\
    \midrule
    \textbf{Average} & $\mathbf{0.0016 \pm 0.0008}$ & $\mathbf{0.0021 \pm 0.0010}$ & $\mathbf{12.61 \pm 5.04}$ & $\mathbf{91.80 \pm 10.97}$ \\
    \bottomrule
    \end{tabular}
\end{table}

\begin{table}[htbp]
\centering
\scriptsize
\caption{Error Metrics Across Measurement Positions for FExD model}
\label{tab:full_err_FExD}
\begin{tabular}{l|r|r|r|r}
\toprule
Position & MAE (m/s$^2$) & RMSE (m/s$^2$) & RRMSE (\%) & R$^2$ (\%) \\
\midrule
1 & $0.0012 \pm 0.0003$ & $0.0016 \pm 0.0004$ & $7.95 \pm 2.91$ & $99.28 \pm 0.68$ \\
2 & $0.0014 \pm 0.0004$ & $0.0018 \pm 0.0005$ & $4.97 \pm 1.46$ & $99.73 \pm 0.18$ \\
3 & $0.0019 \pm 0.0005$ & $0.0025 \pm 0.0007$ & $4.63 \pm 1.42$ & $99.76 \pm 0.18$ \\
4 & $0.0013 \pm 0.0004$ & $0.0017 \pm 0.0005$ & $6.20 \pm 2.38$ & $99.54 \pm 0.40$ \\
5 & $0.0015 \pm 0.0005$ & $0.0019 \pm 0.0006$ & $3.63 \pm 1.23$ & $99.85 \pm 0.12$ \\
6 & $0.0013 \pm 0.0004$ & $0.0017 \pm 0.0006$ & $3.94 \pm 1.82$ & $99.81 \pm 0.21$ \\
7 & $0.0005 \pm 0.0001$ & $0.0007 \pm 0.0001$ & $37.48 \pm 9.50$ & $85.05 \pm 7.81$ \\
8 & $0.0015 \pm 0.0004$ & $0.0019 \pm 0.0005$ & $4.46 \pm 1.59$ & $99.77 \pm 0.19$ \\
9 & $0.0012 \pm 0.0004$ & $0.0016 \pm 0.0005$ & $3.97 \pm 1.36$ & $99.82 \pm 0.14$ \\
10 & $0.0013 \pm 0.0004$ & $0.0016 \pm 0.0005$ & $4.35 \pm 1.32$ & $99.78 \pm 0.15$ \\
11 & $0.0015 \pm 0.0005$ & $0.0020 \pm 0.0006$ & $4.59 \pm 1.60$ & $99.75 \pm 0.20$ \\
12 & $0.0014 \pm 0.0004$ & $0.0018 \pm 0.0004$ & $5.64 \pm 1.77$ & $99.64 \pm 0.27$ \\
13 & $0.0007 \pm 0.0002$ & $0.0009 \pm 0.0002$ & $7.83 \pm 3.53$ & $99.26 \pm 0.79$ \\
14 & $0.0013 \pm 0.0004$ & $0.0017 \pm 0.0005$ & $5.00 \pm 1.62$ & $99.72 \pm 0.22$ \\
15 & $0.0014 \pm 0.0004$ & $0.0019 \pm 0.0006$ & $4.60 \pm 1.45$ & $99.76 \pm 0.17$ \\
16 & $0.0013 \pm 0.0004$ & $0.0016 \pm 0.0005$ & $3.93 \pm 1.31$ & $99.82 \pm 0.14$ \\
17 & $0.0014 \pm 0.0004$ & $0.0018 \pm 0.0005$ & $4.20 \pm 1.26$ & $99.80 \pm 0.13$ \\
18 & $0.0014 \pm 0.0005$ & $0.0019 \pm 0.0006$ & $3.83 \pm 1.33$ & $99.84 \pm 0.13$ \\
19 & $0.0003 \pm 0.0001$ & $0.0004 \pm 0.0001$ & $45.78 \pm 11.77$ & $77.65 \pm 11.62$ \\
20 & $0.0013 \pm 0.0004$ & $0.0017 \pm 0.0005$ & $4.04 \pm 1.68$ & $99.81 \pm 0.19$ \\
21 & $0.0017 \pm 0.0005$ & $0.0022 \pm 0.0007$ & $3.69 \pm 1.20$ & $99.85 \pm 0.11$ \\
22 & $0.0014 \pm 0.0004$ & $0.0018 \pm 0.0005$ & $5.38 \pm 1.69$ & $99.67 \pm 0.23$ \\
23 & $0.0018 \pm 0.0005$ & $0.0023 \pm 0.0007$ & $4.04 \pm 1.25$ & $99.82 \pm 0.13$ \\
24 & $0.0014 \pm 0.0004$ & $0.0018 \pm 0.0005$ & $4.13 \pm 1.19$ & $99.81 \pm 0.12$ \\
25 & $0.0014 \pm 0.0003$ & $0.0018 \pm 0.0004$ & $6.66 \pm 1.97$ & $99.52 \pm 0.35$ \\
26 & $0.0008 \pm 0.0002$ & $0.0010 \pm 0.0003$ & $7.00 \pm 3.39$ & $99.40 \pm 1.05$ \\
\midrule
\textbf{Average} & $\mathbf{0.0013 \pm 0.0004}$ & $\mathbf{0.0017 \pm 0.0005}$ & $\mathbf{7.77 \pm 2.42}$ & $\mathbf{98.29 \pm 1.00}$ \\
\bottomrule
\end{tabular}
\end{table}

\end{document}